\documentclass[sigconf]{acmart}

\usepackage{courier} 
\usepackage{graphicx} 
\usepackage{caption} 
\usepackage{amsmath}
\usepackage{amsfonts}

\usepackage{amssymb}
\usepackage{amsthm}
\usepackage{multirow}
\usepackage{makecell}
\usepackage{booktabs}
\usepackage{xcolor}
\usepackage{enumitem}
\usepackage{verbatim}
\usepackage[ruled]{algorithm2e}
\usepackage{subfigure} 
\usepackage{ragged2e}
\usepackage{caption}
\usepackage{appendix}
\usepackage{bm}
\usepackage{longtable}
\usepackage{adjustbox}

\usepackage{microtype}
\usepackage{setspace}
\usepackage{hyperref}
\usepackage{float}
\usepackage{balance}

\newcommand{\eref}[1]{{\ref{#1}}}

\AtBeginDocument{%
  }

\setcopyright{acmlicensed}
\copyrightyear{2025}
\acmYear{2025}

\acmConference[KDD '25]{Proceedings of the 31st ACM SIGKDD Conference on Knowledge Discovery and Data Mining V.1}{August 3--7, 2025}{Toronto, ON, Canada}
\acmBooktitle{Proceedings of the 31st ACM SIGKDD Conference on Knowledge Discovery and Data Mining V.1 (KDD '25), August 3--7, 2025, Toronto, ON, Canada}
\acmDOI{10.1145/3690624.3709244}
\acmISBN{979-8-4007-1245-6/25/08}

\begin{document}

\title{Conservation-informed Graph Learning for Spatiotemporal Dynamics Prediction}



\author{Yuan Mi}
\affiliation{%
  \institution{Renmin University of China}
  \city{Beijing}
  \country{China}}
\email{miyuan@ruc.edu.cn}

\author{Pu Ren}
\affiliation{%
  \institution{Northeastern University}
  \city{Boston}
  \state{Massachusetts}
  \country{USA}}
\email{ren.pu@northeastern.edu}

\author{Hongteng Xu}
\affiliation{%
  \institution{Renmin University of China}
  \city{Beijing}
  \country{China}}
\email{hongtengxu@ruc.edu.cn}

\author{Hongsheng Liu}
\affiliation{%
  \institution{Huawei Technologies}
  \city{Shenzhen}
  \country{China}}
\email{liuhongsheng4@huawei.com}

\author{Zidong Wang}
\affiliation{%
  \institution{Huawei Technologies}
  \city{Shenzhen}
  \country{China}}
\email{wang1@huawei.com}

\author{Yike Guo}
\affiliation{%
  \institution{HKUST}
  \city{Hong Kong}
  \country{China}}
\email{yikeguo@ust.hk}

\author{Ji-Rong Wen}
\affiliation{%
  \institution{Renmin University of China}
  \city{Beijing}
  \country{China}}
\email{jrwen@ruc.edu.cn}

\author{Hao Sun}
\authornote{Corresponding authors.}
\affiliation{%
  \institution{Renmin University of China}
  \city{Beijing}
  \country{China}}
\email{haosun@ruc.edu.cn}

\author{Yang Liu}
\authornotemark[1]
\affiliation{%
  \institution{University of Chinese Academy of Sciences}
  \city{Beijing}
  \country{China}}
\email{liuyang22@ucas.ac.cn}

\renewcommand{\shortauthors}{Yuan Mi et al.}

\begin{abstract}
  Data-centric methods have shown great potential in understanding and predicting spatiotemporal dynamics, enabling better design and control of the object system. However, deep learning models often lack interpretability, fail to obey intrinsic physics, and struggle to cope with the various domains. While geometry-based methods, e.g., graph neural networks (GNNs), have been proposed to further tackle these challenges, they still need to find the implicit physical laws from large datasets and rely excessively on rich labeled data. In this paper, we herein introduce the conservation-informed GNN (CiGNN), an end-to-end explainable learning framework, to learn spatiotemporal dynamics based on limited training data. The network is designed to conform to the general conservation law via symmetry, where conservative and non-conservative information passes over a multiscale space enhanced by a latent temporal marching strategy. The efficacy of our model has been verified in various spatiotemporal systems based on synthetic and real-world datasets, showing superiority over baseline models. Results demonstrate that CiGNN exhibits remarkable accuracy and generalizability, and is readily applicable to learning for prediction of various spatiotemporal dynamics in a spatial domain with complex geometry. 

\end{abstract}




\maketitle

\vspace{-2pt}
\section{Introduction}\label{Introduction}
The principled and accurate modeling and simulation of spatiotemporal dynamical systems play an important role in many science and engineering applications, e.g., physics, meteorology, ecology, social science, material science, etc. 
Classical approaches are primarily rooted in the use of well-posed physics-based models, governed by a system of partial differential equations (PDEs) under certain initial conditions (ICs) and boundary conditions (BCs), where the solutions could be achieved by a properly chosen numerical solver \cite{ames2014numerical}. 
Nevertheless, the complexity of real-world problems poses grand challenges that traditional physics-based techniques struggle to address, especially when the physical \textit{prior} knowledge is incomplete or repeated forward analyses are required to assimilate data. 

The ever-growing availability of data opens up a new avenue to tackle these challenges, resulting in data-centric methods leveraging the power of machine learning. 
Tremendous efforts have been recently placed on developing statistical or deep learning approaches for modeling \cite{brunton2020machine, kochkov2021machine, pan2023neural}, predicting \cite{regazzoni2019machine, ravuri2021skilful, bi2023accurate, zhang2023skilful}, discovering \cite{schmidt2009distilling, brunton2016discovering, floryan2022data}, and controlling \cite{baggio2021data, zhai2023model} the complex behavior of nonlinear dynamics. 
Moreover, graph neural networks (GNNs) \cite{wu2020comprehensive,zhang2020deep}, which integrate topological structures with an inherent connection to the physical world, are uniquely positioned to address above issues. 
Above all methods learn representations from, but highly rely on, rich labeled datasets, which are often expensive to acquire from experiments or simulations in most scientific and engineering problems. 
Due to the model's over-parameterized black-box nature, issues arise in the context of explainability, over-fitting, and generalizability. 
Embedding the \textit{prior} knowledge to form physics-informed learning has shown the potential to alleviate these fundamental issues \cite{karniadakis2021physics}, producing more robust models for reliable prediction \cite{alberts2023physics}.  

To this end, we propose an end-to-end explainable learning framework, aka conservation-informed GNN (CiGNN), to learn complex spatiotemporal dynamics based on limited training data. 
The network is designed to conform to the general differential form of conservation law with a source term for spatiotemporal dynamics, essentially acting as a \textit{prior} knowledge-based inductive bias to the learning process. 
Due to the existence of source terms (e.g., internal reaction, damping, external input), the system might be subjected to entropy inequality leading to a non-conservative state. 
Inspired by Noether's theorem, we depict the conservation property and entropy inequality via the fundamental principle of symmetry \cite{sarlet1981generalizations}, and then propose a symmetry-preserved MPNN (SymMPNN) module for feature representation learning which accounts for symmetric and asymmetric information passing in graph edges. 
The learning process is resorted to a multi-mesh graph which enables the network to capture local and non-local spatial dependency, facilitating the acquisition of more detailed information across scales. 
To improve the stability and accuracy of long-term rollout prediction, we design a latent time integrator in the network as a time marching strategy which models the temporal dependency between sequenced latent embeddings extracted from each SymMPNN layer. 
The efficacy of CiGNN has been demonstrated on both synthetic and real-world observed datasets for modeling various types of spatiotemporal dynamics within different domain geometric configurations.
The primary contributions can be summarized as follows:
\begin{itemize}
    \item We design a novel model to preserve the general conservation law based on the fundamental principle of symmetry.
    \item Our proposed spatial and temporal schemes effectively control the error accumulation for long-term prediction.
    \item The CiGNN exhibits remarkable accuracy with limited training data (e.g., a few or dozens of datasets), and possesses generalization capability (over ICs, BCs and geometry meshes) in learning complex spatiotemporal dynamics, showing superiority over several representative baselines. 
\end{itemize}

\section{Related Works}\label{Related Works}
\begin{figure*}[t!]
\centering
\includegraphics[width=0.9\textwidth]{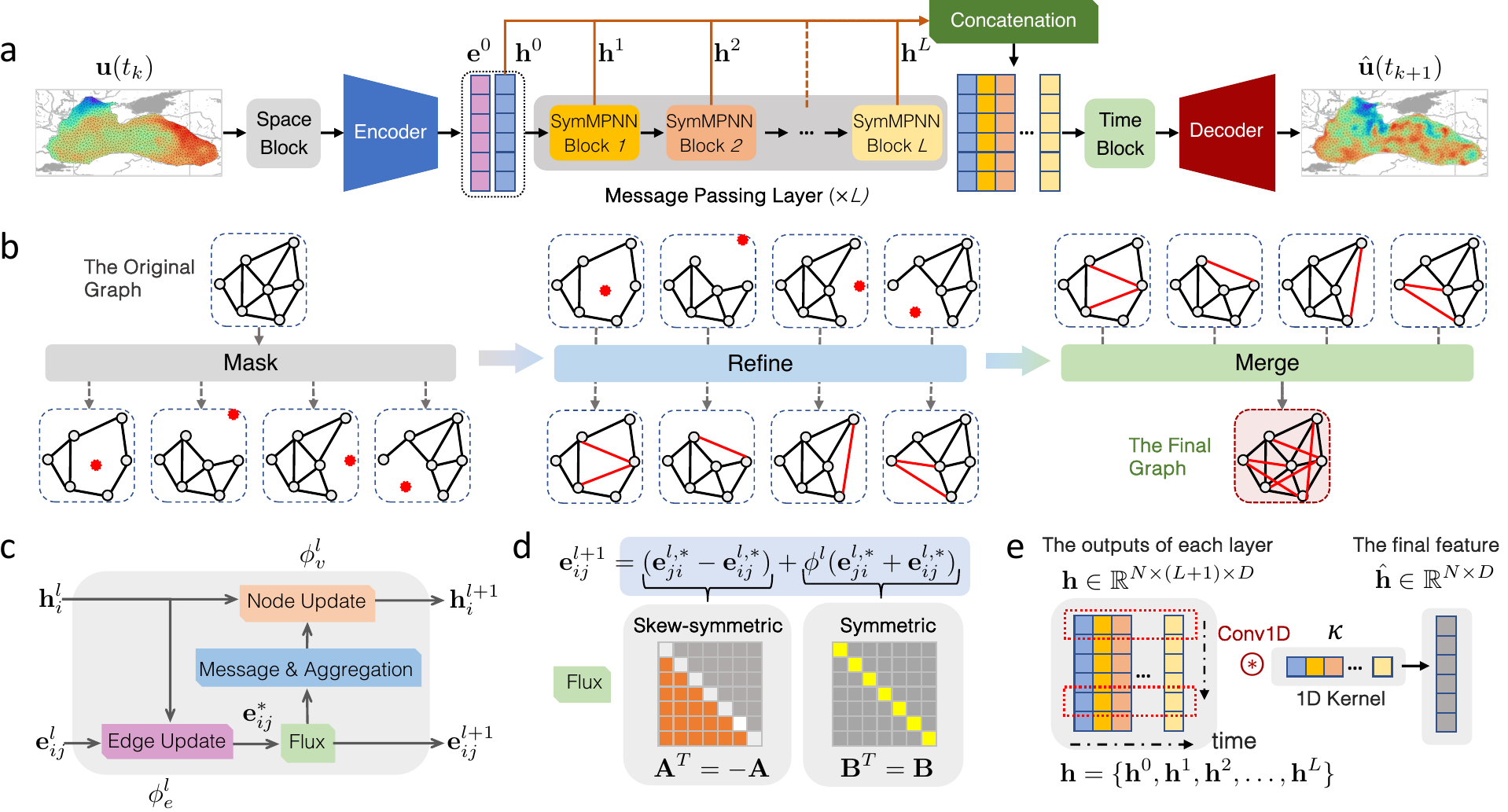}
\vspace{-6pt}
\caption{{Schematic of the CiGNN model.}
\textbf{a}, The main network architecture.
\textbf{b}, The space block that defines a multi-mesh graph to account for different scales and transforms the physical state to a low-dimensional representation. 
\textbf{c}, The SymMPNN block that is designed for learning high-dimensional representation. 
\textbf{d}, The Flux update component that passes symmetric and asymmetric information on edges. 
\textbf{e}, The time block that marches the sequenced high-dimensional features. 
}
\Description{Schematic of the conservation-informed graph neural network (CiGNN) model}
\label{fig:architecture}
\vspace{-6pt}
\end{figure*}

Over the past few decades, classical numerical methods \cite{eymard2000finite,hirani2003discrete} were dominated in the forward and inverse physics problems. However, the development of this field reaches a bottleneck until the deep learning methods' emergence, which changes the status quo.


\vspace{-6pt}
\paragraph{\textbf{Physics-informed learning}}
As one of the most popular framework, the seminal development of the physics-informed neural network (PINN) \cite{raissi2019physics,jagtap2020conservative} has enabled learning in small data regimes for forward and inverse analyses of PDEs, where the loss function is regularized by the PDE residual.
Such a differentiable paradigm has kindled enthusiastic attention in the past few years, which, together with its variant schemes, has been demonstrated effective in modeling and discovering a wide range of complex systems including fluid flows \cite{raissi2020hidden}, cardiovascular systems \cite{arzani2021uncovering}, solid continua's deformation \cite{niu2023modeling}, material constitutive relations \cite{rezaei2024learning}, governing equation discovery \cite{chen2021physics}, data superresolution \cite{ren2023physr}, elastic waves \cite{rao2021physics, ren2024seismicnet}, structural metamodeling \cite{zhang2020physics1, zhang2020physics2}, among many others. 
However, PINN generally has the need of explicit equations and the issue of scalability, resorting to the use of fully-connected neural networks and mesh-free point-wise training. 

\vspace{-6pt}
\paragraph{\textbf{Neural operators}}
Another framework, neural operator, is designed to learn the underlying Green's functions of PDEs, offering an alternative to model spatiotemporal dynamics, which possess generalizability and do not require \textit{a prior} knowledge of the explicit PDE formulation \cite{wu2022learning,wu2024uncertainty}. 
Neural operators, including DeepONet \cite{lu2021learning}, Fourier neural operator (FNO) \cite{li2021Fourier}, and their variant algorithms \cite{tran2023factorized, lanthaler2023nonlinear, wangp2024}, have taken remarkable leaps in approximating the solution of PDE systems. 
However, the generalizability of these models largely depends on the richness of the labeled training data. When the complexity of the considered system remains high (e.g., multi-dimensional spatiotemporal dynamics in an irregular domain geometry), these methods might suffer from scalability issues with deteriorated performance.

\vspace{-6pt}
\paragraph{\textbf{Geometric learning}} 
Recently, geometric learning of system dynamics has drawn great attention \cite{pineda2023geometric,hao2023gnot,wang2024beno}, where graphs, regardless of Euclidean or non-Euclidean meshes, are employed to represent physical features and provide naturally structured relationships (e.g., spatial proximity) with interpretability \cite{liu2024segno,wu2024equivariant}.
For example, message-passing neural networks (MPNNs) have been regarded as an effective form of graph neural network (GNN) in simulating dynamics of particle assembly \cite{sanchez2020learning}, continuum media \cite{pfaff2021Learning} and rigid bodies \cite{han2022learning}, as well as forecasting global weather \cite{lam2023learning}. 
In particular, when the explicit PDE formulation is completely unknown or partially missing, a recent study \cite{hernandez2023thermodynamics} shows that forcibly encoding the law of thermodynamics into the GNN architecture can boost the generalizability of the model for simulation of quasi-static deformable bodies. 
Moreover, the model in \cite{grega2024energy} is designed to deal with lattice meshes and the graph solver in \cite{horie2024graph} is rooted in the assumption of conservative systems to update the discretized conservation equation in the latent space via hard encoding.

\section{Preliminary and Problem Statement}\label{Preliminary and Problem Statement}
Learning to predict the dynamics of complex spatiotemporal systems based on limited training data (e.g., sparsely, and perhaps non-uniformly, sampled in space and time) is regarded as a great challenge \cite{boulle2023elliptic}. 
Our objective is to tackle this challenge via creating a new explainable and generalizable learning framework adaptable to a broad category of spatiotemporal dynamics. 
Drawing from the fundamental insights of physics, we recognize the existence of numerous conservation laws within the physical realm.
In general, the strong form of these conservation laws necessitates that, in order for a conserved quantity at a specific point to alter, there must exist a flow or flux of that quantity either into or out of that point. 
For example, the conservation law with a source term for spatiotemporal dynamics can be presented through a continuity equation, given by
\begin{equation}
\label{eq:continuity}
\frac{\partial \mathbf{u}}{\partial t} + \boldsymbol{\nabla} \cdot \boldsymbol{\mathcal{F}}(\mathbf{u}) = \mathbf{s}(\mathbf{u}; \mathbf{x}, t)
\end{equation}
where $\mathbf{u}(\mathbf{x}, t)\in\mathbb{R}^d$ denotes the system state variable vector composed of $d$ components, $\mathbf{x}\in\mathbb{R}^m$ the $m$-dimensional spatial coordinate, $t$ the time, $\boldsymbol{\nabla}$ the Nabla operator, $\boldsymbol{\mathcal{F}}(\cdot)\in\mathbb{R}^{m\times d}$ the flux function, and $\mathbf{s}(\cdot)\in\mathbb{R}^d$ the source term. 
This equation models the evolution of the system state $\mathbf{u}(\mathbf{x},t)$ over time as the composition of local flux term and the source term.

Now we consider an unstructured mesh of a $m$-dimensional domain (regardless of regular or irregular geometry), where the system state measured at a generic discrete time step $t_{k}$ ($k=1,2,...,T^{'}$) is denoted by $\mathbf{u}(t_{k})\in\mathbb{R}^{N\times d}$. Here, $N$ represents the number of graph nodes (observation grid points) and $T^{'}$ the number of time steps. Our aim is to establish a model to learn spatiotemporal dynamics on a predefined graph $G=(V,E,w)$, which underlines the intrinsic mapping between the input state variable $\mathbf{u}(t_{k})$ and the output subsequent $s$ time series $\{\hat{\mathbf{u}}(t_{k+1}), \dots, \hat{\mathbf{u}}(t_{k+s})\}$. Here, $V$ is a set of nodes, $E$ a set of edges connecting the nodes $V$, and $w(\cdot, \cdot): E \rightarrow \mathbb{R}^{1}$ the edge weight function. For simplicity, $w(\mathbf{u}_{i},\mathbf{u}_{j})$ is expressed by $w_{ij}$ and $w(\mathbf{u}_{j},\mathbf{u}_{i})$ by $w_{ji}$. In the same way, this simplicity also applies to the edge update function $F(\cdot,\cdot)$.

\paragraph{\textbf{Conservation on graph}} Given on the theory of graph calculus, the general conservation law in Eq. \ref{eq:continuity} can be reformulated into a discrete form on graph (see Appendix Section \ref{si:conservative_property_of_CiGNN} for detailed derivation), shown as follows.
\begin{equation}
\label{eq:graph divergence operator on directed graph}
\frac{\partial \mathbf{u}_i}{\partial t} = \frac{1}{2} \sum_{j \in \mathcal{N}_{i}}  \left(\sqrt{w_{ji}}F_{ji}-\sqrt{w_{ij}}F_{ij} \right) + {s}(\mathbf{u}_{i}).
\end{equation}

We realize that the conservation term can be formulated by the skew-symmetric tensor and the inequality by the symmetric tensor \cite{tadmor1984skew}. Instead of constructing such tensors directly, we implicitly build a learnable scheme to retain the above structure property, namely,
\begin{equation}
\label{eq:CiGNN_in_body}
    \frac{\partial \mathbf{u}_i}{\partial t} = 
    \sum_{j \in \mathcal{N}_{i}}  
    \Big(
    \underbrace{ F_{ji}-F_{ij}}_{\text{skew-symmetric}} 
    + 
    \underbrace{ \phi\left(F_{ji}+F_{ij}\right)}_{\text{symmetric}} 
    \Big),
\end{equation}
where $\phi(\cdot)$ is a differential function to model the inequality and, meanwhile, learn the unknown source term (e.g., entropy inequality). 
Here, the skew-symmetric and symmetric features on the right-hand side of Eq.~\eref{eq:CiGNN_in_body} are constructed to approximate the conservation and inequality principles, respectively.
Meanwhile, we omit the coefficient of 1/2 for convenience. 

\section{Methodology}\label{Methodology}
We develop a learning scheme that conforms to the conservation law shown in Eq. \eref{eq:continuity}. 
The schematic overview of our proposed CiGNN is shown in Figure \ref{fig:architecture} which illustrates an end-to-end learning pipeline of ``Encoder-Processor-Decoder'' framework. 
We have released the source code and data at \textit{\url{https://github.com/intell-sci-comput/CiGNN}}.

\subsection{Conservation-informed graph neural network}\label{sub:cignn}

The CiGNN architecture is composed of five key blocks, as shown in Figure \ref{fig:architecture}\textit{a}, including the space block, the encoder, the SymMPNN block, the time block, and the decoder. 
Here, the encoder lifts the low-dimensional quantities to a high-dimensional representation, and in the meantime the decoder projects the updated high-dimensional latent features to the evolved physical state. 
The space block (see Figure \ref{fig:architecture}\textit{b}) defines a multi-mesh graph to account for different scales and transforms the physical system state to a low-dimensional graph representation (e.g., node features, graph topology). 
Given the fact that the disturbance of existing source, such as internal reaction, damping or external input, breaks down the conservation, the system will be subjected to entropy inequality. 
We introduce a multi-layer SymMPNN module (see Figure \ref{fig:architecture}\textit{c--d}) as the core component for high-dimensional feature representation learning, where we construct symmetric and asymmetric matrices to pass non-conservative information in graph edges. 
In the next work, instead of treating the SymMPNN module as a step-by-step learning and acquisition of higher-order information in space, we view the multi-layer message-passing mechanism of SymMPNN as a process of cascaded transmission over time. 
This implies that implicit latent time intervals, either fixed or non-fixed, can be generated between the SymMPNN layers. 
Hence, we propose a simple but effective latent time integrator (see the time block depicted in Figure \ref{fig:architecture}\textit{e}) to march the sequenced high-dimensional features, which improves the stability and accuracy of long-range multi-step rollout prediction.
Moreover, we provide a pseudo-code describing our proposed graph operator, as shown in Algorithm \ref{alg:algorithm1}.

\subsection{Network architecture}\label{sub:network_architecture}


\subsubsection{Encoder}

The Encoder lifts the physical state into latent features represented in a high-dimensional space. 
The encoded initial node features $\mathbf{h}_{i}^{0}\in\mathbb{R}^c$ imply the physical state  $\mathbf{u}_i$ at the $i$th node, the corresponding position feature $\mathbf{x}_{i}$ and the node type, where $c$ denotes the channel size (aka, the dimension of the latent space). 
The encoded initial edge features $\mathbf{e}_{ij}^{0}\in\mathbb{R}^c$ capture the relative positional information between the connected $i$th and $j$th nodes. 
Additionally, a variety of variables, such as angles, are encoded to enhance the expressiveness of the edge feature representation. 
The corresponding formulations are given by 
\begin{subequations}
\begin{align}
\mathbf{h}_{i}^{0}&=\phi_{v}^{en} \left(\mathbf{u}_{i} \parallel \mathbf{x}_{i} \parallel \varkappa_{i} \parallel \dots \right),\\
\mathbf{e}_{ij}^{0}&=\phi_{e}^{en} \left(\left(\mathbf{x}_{j}-\mathbf{x}_{i}\right) \parallel d_{ij} \parallel \vartheta_{ij}^x \parallel \vartheta_{ij}^y \parallel \dots \right),
\end{align}
\end{subequations}
where $\phi_{v}^{en}(\cdot)$ and $\phi_{e}^{en}(\cdot)$ represent the learnable functions (e.g., MLPs); $\mathbf{x}_{j}-\mathbf{x}_{i}$ a vector that reflects the relative information between the $i$th and $j$th nodes; $d_{ij}$ the relative distance in the physical space; $\varkappa_{i}$ the $i$th node type; and $\vartheta_{ij}^x, \vartheta_{ij}^y$ the angle information of edge in different directions. 
In addition, $(\cdot \parallel \cdot)$ denotes the concatenation.

\begin{algorithm}[t!]
\caption{Conservation-informed Graph Operator}
\label{alg:algorithm1}
\KwIn{The current states $ \mathbf{g}^{l}, \mathbf{h}_i^{l}, \mathbf{e}_{ij}^{l} $.}
\KwOut{The next states $ \mathbf{g}^{l+1}, \mathbf{h}^{l+1}_i, \mathbf{e}^{l+1}_{ij} $.}
\While{stop condition is not reached}{
\For{each edge $\mathbf{e}_{ij}^{l}$}
{ 
Gather sender and receiver nodes $\mathbf{h}_{i}^{l}, \mathbf{h}_{j}^{l}$\;
Compute output edge
$\mathbf{e}^{l,*}_{ij} = \phi_{e}^{l} \left(\mathbf{g}^{l}, \mathbf{h}_{i}^{l}, \mathbf{h}_{j}^{l}, \mathbf{e}_{ij}^{l} \right)$\;
}
\For{each edge $\mathbf{e}_{ij}^{l,*}$}
{
Obtain the reversed edge $\mathbf{e}^{l,*}_{ji}$ with the edge $\mathbf{e}^{l,*}_{ij}$\;
Get the new edge state $\mathbf{e}_{ij}^{l+1}$ with the flux function: 
$\mathbf{e}^{l+1}_{ij} = \left(\mathbf{e}^{l,*}_{ji} - \mathbf{e}^{l,*}_{ij} \right) + \phi^{l} \left(\mathbf{e}^{l,*}_{ij} + \mathbf{e}^{l,*}_{ji} \right)$\;
}
\For{each node $\mathbf{h}_i^{l}$}
{ 
Aggregate $\mathbf{e}^{l+1}_{ij}$ of per receiver, get the sum of the edge features $\bar{\mathbf{e}}_{i}^{l+1} = \sum\limits_{j \in \mathcal{N}_{i}} \mathbf{e}^{l+1}_{ij}$\;
Compute the new node-wise feature $\mathbf{h}^{l+1}_i = \phi_{v}^{l} \left(\mathbf{g}^{l}, \mathbf{h}_i^{l}, \bar{\mathbf{e}}_{i}^{l+1} \right)$\;
}
Aggregate all edges and nodes $\tilde{\mathbf{e}}^{l+1} = \sum^{N} \limits_{i} \bar{\mathbf{e}}^{l+1}_{i}$, $\tilde{\mathbf{h}}^{l+1} = \sum^{N}\limits_{i} \mathbf{h}^{l+1}_i$\;
Compute new global features $\mathbf{g}^{l+1} =  f_{g} \left(\mathbf{g}^{l}, \tilde{\mathbf{h}}^{l+1}, \tilde{\mathbf{e}}^{l+1} \right)$\;
}
return the next states $ \mathbf{g}^{l+1}, \mathbf{h}^{l+1}_i, \mathbf{e}^{l+1}_{ij} $. 
\end{algorithm}

\subsubsection{Processor}

The Processor employs a graph with exclusively inter-node connections, where spatial and channel information from the local neighborhood of each node is aggregated through connections to adjacent nodes~\cite{sanchez2020learning}. 
As shown in Figure \ref{fig:architecture}\textit{a}, it iteratively processes the high-dimensional latent features using a sequence of SymMPNN layers. 
For each layer, each node exchanges information with itself and its neighbors. 
Furthermore, the architecture incorporates residual connections between each processing layer, which allow for direct message passing at different scales. 
This part introduces the core component of the architecture, the conservation-informed graph operator (see \textcolor{black}{Algorithm 1}). The design of information passing across nodes and edges is inspired by the conservation on graph as shown in Eq. \ref{eq:CiGNN_in_body}. 
Given a graph with available node and edge features, the mathematical formulation for the message-passing mechanism at each SymMPNN layer is defined as 
\begin{subequations}
\begin{align}
\mathbf{e}^{l,*}_{ij} &=\phi_{e}^{l} \left(\mathbf{h}_{i}^{l}\parallel\mathbf{h}_{j}^{l}\parallel\mathbf{e}_{ij}^{l} \right ), \label{edge1} \\
\mathbf{e}^{l+1}_{ij} &=  \underbrace{\left(\mathbf{e}^{l,*}_{ji} - \mathbf{w}^l\circ\mathbf{e}^{l,*}_{ij} \right)}_{\text{asymmetric}} + \underbrace{\phi^l \left(\mathbf{e}^{l,*}_{ji}+\mathbf{e}^{l,*}_{ij} \right)}_{\text{symmetric}}, \label{edge2} \\
\mathbf{h}_{i}^{l+1}&=\phi_{v}^{l} \bigg(\mathbf{h}_{i}^{l}\parallel\sum_{j \in \mathcal{N}_{i}} \mathbf{e}^{l+1}_{ij} \bigg), \label{node}
\end{align}
\end{subequations}
where $\mathbf{h}_{i}^{l}\in\mathbb{R}^c$ and $\mathbf{h}_{j}^{l}\in\mathbb{R}^c$ denote the high-dimensional node features at nodes $i$ and $j$, respectively, at the $l$th SymMPNN layer; $\mathbf{e}_{ij}^{l}\in\mathbb{R}^c$ the high-dimensional edge features between the $i$th and $j$th nodes; $\mathbf{e}^{l,*}_{ij}\in\mathbb{R}^c$ and $\mathbf{e}^{l,*}_{ji}\in\mathbb{R}^c$ the corresponding intermediate edge features (e.g., bidirectional) inferred to represent symmetric and asymmetric information passing; $\phi_{e}^{l}(\cdot)$ and $\phi_{v}^{l}(\cdot)$ the edge and node update functions (e.g., MLPs); $\mathbf{w}^l\in\mathbb{R}^c$ the trainable weights shared over the entire graph for the $l$th layer; $\mathcal{N}_{i}$ the neighborhood node's index set for node $i$; and $\circ$ the Hadamard product. Here, $\mathbf{e}^{l+1}_{ij}$ facilitates the learning process by strictly obeying the conservation law and the inequality principle. $\phi^l(\cdot)$ a flux update function (e.g., MLP) for symmetric message passing. 


We found that, when $\mathbf{w}^l\equiv 1$, our CiGNN model achieves a more reliable performance. 
For comparison purpose, we remove the symmetric update part as shown in Eq. \eref{edge2} while only keeping the asymmetric portion. Thus, we obtain two variant models, namely, CiGNN$^-$ ($\mathbf{w}^l\equiv 1$) and CiGNN$^*$ ($\mathbf{w}^l\neq 1$, which is trainable).

\subsubsection{Decoder}

The Decoder maps updated high-dimensional latent features back to the original physical domain, which receives input from the Processor and generates the predicted increment for the last time step as output. 
This increment is then added as residue to the initial state to determine the new state. The learning process at the Decoder is given by 
\begin{equation}
\hat{\mathbf{u}}_{i} \left (t+\Delta t \right )=\phi_{v}^{de} \left(\mathbf{h}_{i}^{L} \right)+\mathbf{u}_{i} \left(t\right),
\end{equation}
where $\phi_{v}^{de}(\cdot)$ is a differentiable function (e.g., MLP) and $L$ the total number of SymMPNN layers.

\vspace{0pt}
\subsection{Spatial and temporal learning strategies}\label{sub:spatial_and_temporal_learning_strategy}

\subsubsection{Space Block with multi-mesh strategy}

Consider a graph with $G=(V,E)$, where $V$ denotes the node set and $E$ the edge set. 
The multi-mesh strategy involves three steps: (1) defining $G^{k}$ as the initial graph; (2) randomly discarding a proportion of nodes in $G^{k}$ to generate a new graph $G^{k+1}$ as the subsequent initial graph; (3) iteratively repeating the second step, and aggregating all graphs into $G$. 
We organize the generated graphs by layers, designating the graph with the most node count as the topmost layer and the one with the fewest nodes as the bottom layer. In other words, as $k$ increases ($k=0,1,2,...$), the resulting graph structure $G^{k+1}$ contains fewer nodes.
Then, we define a refinement ratio $r \in [0, 1]$ to characterize the relationship between node numbers in the context graph. For instance, if $G^{k}$ has $N$ nodes, then the subsequent graph, $G^{k+1}$, will have $rN$ nodes. We specifically set the refinement ratio $r$ to 0.1 in this study. 
The overall graph $G$ can be conceptualized as the union of all edges and nodes:
\begin{equation}
G = \bigcup_{k} G^{k},~~~~G^{k}= \left(V^{k},E^{k} \right),~~~~k=0,\dots,m.
\end{equation}
Noting that the nodes in $G^{k+1}$ are always the node subset of $G^{k}$, and the edge reconstruction process can capture information across extended distances. This leads to a multi-scale mesh distribution, enabling the network to capture both local and non-local relationships and learn complex detailed information. Meanwhile, the multi-mesh technique performs edge reconstruction only during pre-processing (e.g., not training), following the random masking of nodes at each level. As shown in Figure \ref{fig:architecture}\textit{b}, the mask operation generates numerous subgraphs, which can alleviate the over-reliance on local information and enhance node-level feature extraction. Generally, the multi-mesh strategy can facilitate acquiring the key structural information. 

\begin{table*}
\caption{
Summary of each model's performance in terms of prediction error, e.g., aggregated root mean square error (RMSE) between the predicted and ground truth test data.
Here, ``--'' denotes the model is unable or unsuitable to learn the dynamics (e.g., inapplicable to handle irregular mesh). ``MP-PDE$^{*}$'' means that no gradient is backward propagated except for the last step. 
The \textbf{bold} values and \underline{underlined} values represent the optimal and sub-optimal results on various datasets. 
``Pro.'' means the promotion value which is calculated from the optimal and sub-optimal results.}
\vspace{-6pt}
\label{tab:performance summary}
\centering
\begin{tabular}{ccccc}
\toprule
Model    &\makecell[c]{2D Burgers ($\Delta t = 0.001$ s)} & \makecell[c]{3D GS RD  ($\Delta t = 0.001$ s)}& \makecell[c]{2D CF ($\Delta t = 0.01$ s)}& \makecell[c]{2D BS  ($\Delta t = 1 $ day)}\\
\midrule
\multirow{1}{*}{DeepONet} 
       & $3.7752\times 10^{-1}$     & $1.3954\times 10^{1}$   & $9.1095 \times 10^{-1}$  & $2.8696\times 10^0$        \\
\midrule
\multirow{1}{*}{FNO} 
  & $1.6005\times 10^{-2}$     & $1.5184\times 10^{-1}$   & --                       & --         \\
\midrule 
\multirow{1}{*}{MGN} 
  & $1.8955\times 10^{-2}$     & $1.0350\times 10^{-2}$   & $\underline{1.5790\times 10^{-1}}$   & $\underline{6.9512\times 10^{-1}}$ \\
\midrule
\multirow{1}{*}{MP-PDE$^{*}$}  
& $1.6280\times 10^{-2} $    & $\underline{1.0186\times 10^{-2}}$   & $2.4511\times 10^{-1}$   & $9.0051\times 10^{-1}$ \\ 
\midrule
\multirow{1}{*}{MP-PDE} 
& $\underline{1.5551\times 10^{-2}}  $   & $1.0843\times 10^{-2}$   & $1.6412\times 10^{-1}$   & $6.9621\times 10^{-1}$ \\ 
\midrule
\multirow{1}{*}{CiGNN} 
& $\mathbf{4.2916\times 10^{-3}} $                      & $\mathbf{8.3840\times 10^{-3}}$
            & $\mathbf{1.2561\times 10^{-1}} $                      & $\mathbf{6.5503\times 10^{-1}}$ \\
\midrule
\multirow{1}{*}{Pro. (\%) $\uparrow$} &72.4&17.7&20.4&5.7 \\
\bottomrule
\end{tabular}
\vspace{-2pt}
\end{table*}

\subsubsection{Time Block with trainable temporal integration}

We treat the multi-layer message-passing mechanism of SymMPNN as a process of cascaded transmission over time. A series of fixed or non-fixed sub-time intervals $\Delta t^{*}_{j}$ are generated implicitly, e.g., $\Delta t^{*}_{j} \in [0,\Delta t_k]$, where $j = 0, 2, ..., L-1$ and $\Delta t_k$ denotes the physical time step. Given this assumption, each SymMPNN layer can be regarded as the evolution of information over one time interval. This is equivalent to the sub-stepping process commonly used in numerical integration methods. Additionally, a Taylor series expansion shows that the method is consistent if and only if the sum of the coefficients $\alpha_{j}$ is equal to one,  described as
\begin{subequations}
\begin{align}
\mathbf{h}^{L}  &= \mathbf{h}^{0} + \sum_{j=0}^{L-1}\alpha_{j} \mathbf{K}_{j} \quad \text{s.t.}~~\sum_{j=0}^{L-1} \alpha_{j}=1,  \label{coeff1}\\
\mathbf{K}_{j+1} &= \varphi_{j+1} \left (t + \Delta t^{*}_{j}, \mathbf{h}^{0} + \beta_{j} \mathbf{K}_{j} \right), \label{coeff2}
\end{align}
\end{subequations}
where $\alpha_{j}, \beta_{j}$ are scalar, $\mathbf{K}_{0} = \mathbf{h}^{0}$ the initial node features, $\varphi_{j+1}$ the learnable function in the $(j+1)$th SymMPNN block, and $L$ the total number of SymMPNN layers. 

Notably, this temporal integration strategy is realized on the high-dimensional space, so the coefficients $\alpha_{j}, \beta_{j}$ are also learnable latent features.
$\mathbf{K}_{L}$ denotes the output of final SymMPNN layer, which is the $L$-order approximation of solution. Each increment can be viewed as the product of the coefficients $\beta_{j}$ and an estimated slope specified by each function $\varphi_{j}$. The total increment is the sum of the weighted increments. To this end, we employ a trainable 1D convolutional filter to approximate the coefficients in Eqs. \eref{coeff1} and \eref{coeff2}, as shown in Figure \ref{fig:architecture}\textit{d}. This temporal integration strategy does not lead to extra cost for network training and inference.

\begin{figure}[t!]
\centering
\includegraphics[width=0.8\columnwidth]{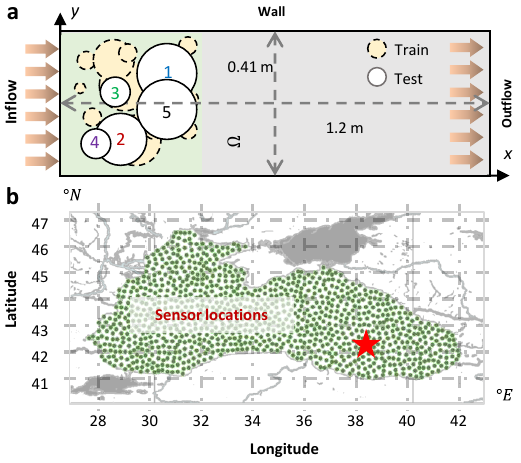}
\vspace{-6pt}
\caption{{Diagrams of irregular domains.}
\textbf{a}, Schematic diagram of the CF example setup with the fluid flowing in from the inflow to the outflow.
\textbf{b}, The geographic and sensor information of the BS Real-world dataset.
}
\Description{Diagrams of irregular domains.}
\label{fig:examples}
\vspace{-6pt}
\end{figure}

\section{Experiments}\label{Experiments}
\begin{table*}
\caption{
{Basic information of datasets.}
} 
\vspace{-6pt}
\label{tab:datasets}
\centering
\begin{tabular}{llllllll}
\toprule
Dataset  & \makecell[l]{Domain} & \makecell[l]{Physical \\ parameters}& \makecell[l]{No. of \\ nodes} &\makecell[l]{Trajectory \\length}&\makecell[l]{Train/Validation/Test \\trajectories}&\makecell[l]{Boundary \\ condition} &  \makecell[l]{Force \\term}\\
\midrule
\makecell[l]{2D Burgers} & Square $[0,1]^{2}$&  ($u,v$) &2,500 ($50^{2}$)&1,000&60 (50/5/5)& Periodic &No\\
\makecell[l]{3D GS RD} & Cubic $[0,96]^{3}$& ($u,v$) & 13,824 ($24^{3}$)& 3,000 &10 (5/3/2)& Periodic &No\\
\makecell[l]{2D CF} & Irregular & ($u,v,p$) & 3,000$\sim$5,000 & 1,000 &220 (200/10/10)& Unknown & No\\
\makecell[l]{2D BS} & Irregular &  ($u,v,T$) &1,000$\sim$40,000&365 &29 (25/3/1)& Unknown & Unknown\\
\bottomrule
\end{tabular}
\vspace{-2pt}
\end{table*}

\vspace{-2pt}
\subsection{Datasets and Baselines.}\label{sub:datasets_and_baselines}

To verify the efficacy of CiGNN for learning complex spatiotemporal dynamics on graphs, we consider four different systems shown in \textcolor{black}{Table \ref{tab:datasets}}, e.g., the 2D Burgers equation, a 3D Gray-Scott (GS) reaction-diffusion (RD) system, the 2D incompressible flow past a cylinder (CF), and the Black Sea (BS) hydrological observation (see Figure \ref{fig:examples}). The datasets of the first three systems were spatiotemporally down-sampled (e.g., 5-fold in time) from high-fidelity data generated by direct numerical simulations considering different ICs, BCs, and meshes, while the last one was compiled from field measurements.
More detailed information of data generation can be found in Appendix Section \ref{si:dataset_generation}.
We selected four baseline models for comparison 
, including deep operator network (DeepONet) \cite{lu2019deeponet}, Fourier neural operator (FNO) \cite{li2021Fourier}, mesh-based graph network (MGN) \cite{pfaff2021Learning}, and message-passing PDE solver (MP-PDE) \cite{brandstetter2022message}.

\vspace{-2pt}
\subsection{Loss function and Evaluation metrics}\label{sub:loss_function_and_evaluation_metrics}

The training procedure is to minimize the discrepancy between the labeled and the predicted data. The loss function is defined as 
\begin{equation}
    \mathcal{L}(\boldsymbol{\theta}) = \frac{1}{dN\widetilde{T}}  \left\|vec\big(\widehat{\mathbf{U}}\big) - vec\big(\mathbf{U}\big) \right\|_2^2,
\end{equation}
which calculates the aggregated mean square errors (MSE) between the rollout-predicted data $\widehat{\mathbf{U}}\in\mathbb{R}^{N\times d\times {T}}$ and the training labels $\mathbf{U}\in\mathbb{R}^{N\times d\times {T}}$ to optimize the trainable parameters ($\boldsymbol{\theta}$) in our model. Here, $vec(\cdot)$ denotes the vectorization operation; $d$ the system dimension; $N$ the total number of mesh grid points; ${T}$ the number of rollout time steps in training.
The root mean squared error (RMSE) and the Pearson correlation coefficient (PCC) are utilized as the evaluation metrics, given by
\begin{subequations}
\begin{align}
RMSE(\mathbf{U}, \hat{\mathbf{U}}) &=  \sqrt{\frac{1}{dNT} \left\|vec(\widehat{\mathbf{U}}) - vec(\mathbf{U})\right\|_2^2 }, \\
PCCs(\mathbf{U}, \hat{\mathbf{U}}) &=  \frac{cov \left[vec(\widehat{\mathbf{U}}), vec(\mathbf{U}) \right]}{\sigma_{\mathbf{u}}\sigma_{\hat{\mathbf{u}}}},
\end{align}
\end{subequations}
where $cov[\cdot, \cdot]$ the covariance function; $\sigma_{\mathbf{u}}$ and $\sigma_{\hat{\mathbf{u}}}$ the standard deviations of $\mathbf{U}$ and $\widehat{\mathbf{U}}$, respectively. In addition, introducing noise to the training data can stabilize the model and improve the generalization ability for learning spatiotemporal dynamics \cite{sanchez2020learning, pfaff2021Learning}. Therefore, we add a small amount of Gaussian noise in the training data to improve the CiGNN's performance.

\subsection{Treatment of Boundary Condition}\label{si:treatment_of_boundary_condition}

We consider hard encoding of BCs to facilitate the network optimization and improve the solution accuracy. For example, the periodic padding technique has demonstrated effective to improve the model's performance \cite{rao2023encoding}. The specific implementations are: (1) labelling the nodes with different types to augment the input features and distinguish the boundary nodes; (2) padding on the boundary nodes with specified values when computing training loss. In particular, we consider two types of BCs, namely, periodic and Dirichlet. For the system with periodic BCs, we augment the graph by adding a set of ghost nodes and conduct periodic padding based on the predicted boundary node features \cite{rao2023encoding}. When the Dirichlet BCs are known, we directly pad the boundary node features with the ground truth boundary values. Note that there is no specific treatment of BCs in the original GNN baseline models, such as MeshGraphNet~\cite{pfaff2021Learning} and MP-PDE~\cite{brandstetter2022message}. To be fair, this method of BC encoding is used as a plug-in for all relevant models.



\begin{figure}[t!]
\centering
\includegraphics[width=\columnwidth]{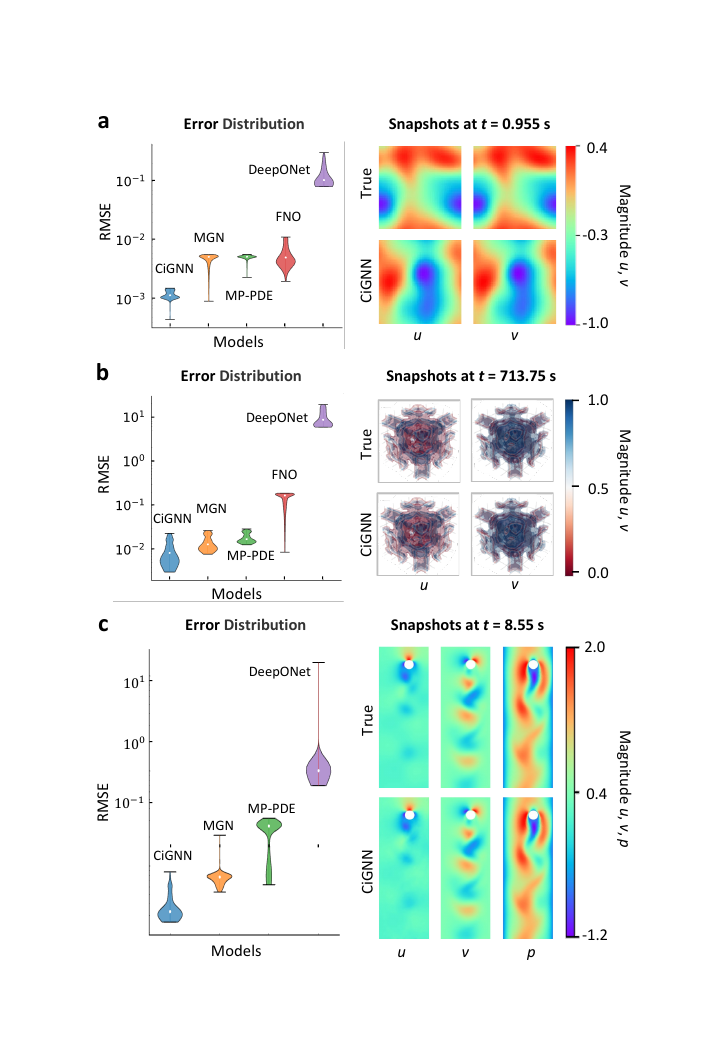}
\vspace{-6pt}
\caption{
{Error distributions and the system state snapshots predicted by CiGNN and other baselines.}
\textbf{a}, The 2D viscous Burgers equation example. 
\textbf{b}, The 3D Gray-Scott equation example. 
\textbf{c}, The 2D CF example setup with the fluid flowing in from the inflow to the outflow. 
}
\Description{Error distributions and the system state snapshots predicted by CiGNN and other baselines.}
\label{fig:mainResult}
\vspace{-20pt}
\end{figure}

\vspace{-2pt}
\subsection{Training settings and resources} \label{sub:training_settings_and_computational_resources}

For a fair comparison, we train each model ten times and select the one with the median performance according to the evaluation of the validation datasets. 
All models are trained with the Adam Optimizer \cite{kingmaB2015adam} and the ReduceLROnPlateau learning scheduler \cite{Paszke2019PyTorch} which decays the learning rate by a factor of $0.8$. 
We train all models for 1,000 epochs and set early stopping with a patience of 100 epochs. 
We utilize the GELU activation function \cite{Paszke2019PyTorch} for all MLP hidden layers, 
 while linear activation is considered for the output layers. 
Additionally, layer normalization is applied after each MLP to improve the training convergence, except the Decoder. 

We use the similar architecture as in \cite{pfaff2021Learning} but only keep the encoder and decoder block. 
For concretely, the encoder module consists of a 2-layer MLP with a hidden size of 128, and the decoder has a 2-layer MLP with a hidden size of 128.
We only use 4 SymMPNN blocks for the processor, where 2 layer MLPs with a hidden size of 128 are employed as the node, edge and flux update functions. The same hyperparameters are used in MGN, MP-PDE, and CiGNN.
Notably, all the hyperparameters are selected from various ranges. For example, the learning rate is selected from the set $[10^{-1},10^{-2},10^{-3},10^{-4},10^{-5}]$.
For hardware, the models were trained on the NVIDIA A100 GPU (80GB) on an Intel(R) Xeon(R) Platinum 8380 CPU (2.30GHz, 64 cores) server.


\vspace{-6pt}
\subsection{Results}\label{sub:results}

\begin{figure}[t!]
\centering
\includegraphics[width=\columnwidth]{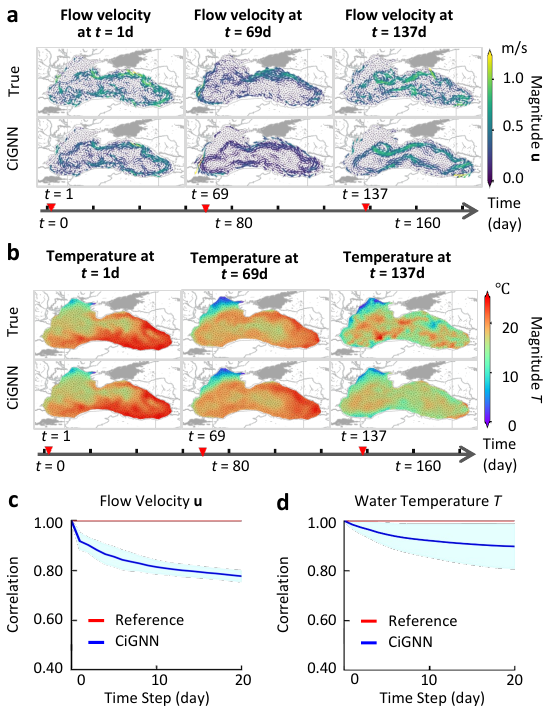}
\vspace{-6pt}
\caption{
{Prediction results of CiGNN on real-world BS hydrological dataset.}
\textbf{a--b}, Snapshots of the predicted flow velocity and water temperature at day 1, day 69, and day 137.
\textbf{c--d}, Pearson correlation of the corresponding predicted flow velocity and water temperature. 
The above three time points represent the sampling moments during the early, middle, and late stages of prediction, respectively.
The blue line represents the mean value of the correlation, and the light blue area the deviation (e.g., $\pm\sigma$).
}
\Description{Prediction results of CiGNN on real-world BS hydrological dataset.}
\label{fig:bs}
\vspace{-6pt}
\end{figure}

\begin{table*}
\caption{
{Ablation results on four datasets.}
Here, the symbol ``--'' represents that Task $\#2$ on CF dataset does not exist since multi-mesh strategy is not utilized.
The \textbf{bold} values and \underline{underlined} values represent the optimal and sub-optimal results.}
\label{tab:Ablation Study}
\vspace{-6pt}
\begin{center}
\begin{tabular}{clcccc}
\toprule
\multirow{1}{*}{Task}&\multirow{1}{*}{Method}& \multicolumn{1}{c}{\makecell[c]{2D Burgers  ($\Delta t = 0.001 s$)}}&\multicolumn{1}{c}{\makecell[c]{3D GS RD  ($\Delta t = 0.001 s$)}} & \multicolumn{1}{c}{\makecell[c]{2D CF   ($\Delta t = 0.01 s$)}}& \multicolumn{1}{c}{\makecell[c]{2D BS  ($\Delta t = 1 day$)}}\\
\midrule 
\#1 &  w/o flux update    & $1.0012\times 10^{-2}$&   $1.3138\times 10^{-2}$  & $1.4526\times 10^{-1}$ &  $6.9581\times 10^{-1}$   \\
\#2  &  w/o multi-mesh       & $9.8175\times 10^{-3}$ &  $1.1101\times 10^{-2}$  & -- &  $6.9456\times 10^{-1}$   \\
\#3 &  w/o temporal         & $\underline{9.2089\times 10^{-3}}$&  $\underline{9.5921\times 10^{-3}}$   & $\underline{1.2740\times 10^{-1}}$ &   $\underline{6.5589\times 10^{-1}}$  \\
\midrule
\#0 & CiGNN (Full)                &  $\mathbf{4.2916\times 10^{-3}}$  &  $\mathbf{8.3840\times 10^{-3}}$   &  $\mathbf{1.2561\times 10^{-1}}$ &   $\mathbf{6.5503\times 10^{-1}}$   \\
\bottomrule
\end{tabular}
\end{center}
\vspace{-6pt}
\end{table*}

\begin{table*}
\caption{
{Summary of performance of CiGNN and its variants.}
CiGNN$^{-}$ and CiGNN$^{*}$ are two variants of CiGNN, where in the SymMPNN block only the asymmetric information passing is considered.
}\label{tab:performance summary of variants}
\vspace{-6pt}
\centering
\begin{tabular}{llcccc}
\toprule
Model   & Rollout Step &2D Burgers ($\Delta t = 0.001 s$) & 3D GS RD ($\Delta t = 0.001 s$)& 2D CF ($\Delta t = 0.01 s$)& 2D BS ($\Delta t = 1 day$)\\
\midrule
\multirow{2}{*}{CiGNN} 
& One       & $4.7870\times 10^{-2}  $   & $9.4229\times 10^{-3}$   & $8.2013\times 10^{-1}$   & $9.2701\times 10^{-1}$ \\
& Multiple  & $\mathbf{4.2916\times 10^{-3}} $                      & $\mathbf{8.3840\times 10^{-3}}$
            & $\mathbf{1.2561\times 10^{-1}} $                      & ${6.5503\times 10^{-1}}$ \\
\midrule
\multirow{2}{*}{CiGNN$^{-}$} 
& One       & $5.0001\times 10^{-2}$     & $1.1274\times 10^{-2}$  & $8.2866\times 10^{-1}$    & $9.0891\times 10^{-1}$ \\ 
& Multiple  & $5.2239\times 10^{-3} $    & $1.0675\times 10^{-2}$  & $1.3678\times 10^{-1}$    & $6.1735\times 10^{-1}$ \\
\midrule
\multirow{2}{*}{CiGNN$^{*}$} 
& One       & $5.6548\times 10^{-2} $    & $9.7794\times 10^{-3}$   & $8.2520\times 10^{-1}$   & $9.0788\times 10^{-1}$ \\ 
& Multiple  &$ 5.6090\times 10^{-3}$     & $9.7826\times 10^{-3}$   & $1.3302\times 10^{-1}$   & $\mathbf{6.1515\times 10^{-1}}$ \\ 
\bottomrule
\end{tabular}
\vspace{-6pt}
\end{table*}

\subsubsection{2D Convection-Diffusion system}

The results listed in Table \ref{tab:performance summary} show that CiGNN consistently outperforms other baseline models in the context of prediction accuracy. 
Figure \ref{fig:mainResult}\textit{a} displays the distribution of prediction errors as well as the predicted snapshots at a typical time step (0.995 s) by different methods. 
The CiGNN model lifts the prediction accuracy by 1$\sim$2 orders of magnitude in terms of MSE metric compared with all other baselines. 
Due to the meshless and point-wise nature of DeepONet, it falls short in scalable modeling of higher dimensional spatiotemporal dynamics. 
Given the squared domain with structured grid mesh, FNO produces a moderate prediction result in the situation with limited training data. 
The performance of MP-PDE aligns with that of MGN.
Note that MP-PDE is a derivative of MGN with the primary distinction in their decoder components, e.g., MGN employs multilayer perceptron (MLP) while MP-PDE utilizes convolution for decoding. 
Despite discrepant performance of other models, CiGNN generalizes well to various ICs and maintains a high prediction accuracy, which demonstrates notable strength and shows clear superiority over all baselines on 2D Burgers equation. 

\vspace{-2pt}
\subsubsection{3D Reaction-Diffusion system}

The results reported in Table \ref{tab:performance summary} once again show our CiGNN model achieves the best performance. 
Figure \ref{fig:mainResult}\textit{b} displays the prediction error distribution and typical snapshots for the GS RD system at 713.75 s. 
Similar to the previous example, FNO, renowned for its potency in the realm of operator learning, appears to inadequately grasp the underlying dynamics of the GS RD equation in the regime of small training data. 
Unfortunately, DeepONet exhibits little progress in learning the intricate evolution of the 3D system states, e.g., the prediction over time still deviates clearly from actual values yielding large errors. 
The results serve to substantiate that both MGN and MP-PDE, based on the multi-step rollout training strategy, maintain robustness. 
However, CiGNN still outperforms other models, showing its superior potential in learning spatiotemporal dynamics in a 3D space. 

\vspace{-3pt}
\subsubsection{2D Flow past a cylinder}

Our model CiGNN generalizes well over $Re$ and BCs as well as graph meshes in the CF example, which again outperforms the baselines as depicted in Table \ref{tab:performance summary}. 
Figure \ref{fig:mainResult}\textit{c} displays the snapshots of the predicted flow field at time 8.55 s in the generalization test, where the cylinder size and position as well as the domain mesh remain different from those in the training datasets. 
Notably, generalizing the learning model on various CF datasets poses grand challenges due to the varying mesh node counts and graph structures, exacerbated by the introduction of random cylinder positions that lead to variant Reynolds numbers. 
Since FNO struggles with irregular mesh structures, we opted not to include it in the comparison experiments. The point-wise meshless learning scheme of DeepONet might suffer from the scalability issue, which, as a result, leads to failure in the CF example. 
The distribution of performance for MGN remains a similar trend illustrated previously. However, MP-PDE shows deterioration compared to the previous two examples. Overall, CiGNN surpasses other baseline models in various degrees. 
Figure \ref{fig:mainResult}\textit{c} shows two additional generalization test results concerning varying domain meshes as well as different locations and sizes (e.g., from small to large) of the cylinder. 
It can be seen from Figure \ref{fig:mainResult}\textit{c} that as the cylinder size decreases, $Re$ increases and the flow becomes more turbulent. Nevertheless, CiGNN can still accurately predict the turbulent flow dynamics. 

\vspace{-2pt}
\subsubsection{2D Hydrological dynamics forecast of the Black Sea}


We employ the trained CiGNN model to conduct a 20-day-long forecasting of the hydrological dynamics of the Black Sea. 
Figure \ref{fig:bs}\textit{a--b} show the results of the predicted water flow velocities and sea surface temperature. It is evident that our model effectively captures the evolution patterns of the hydrological dynamics, especially for low-frequency information that dominates the overall trend (as can be seen from the predicted snapshots). 
The correlation values for both the velocity and temperature fields consistently lie above 0.8, indicating an accurate prediction even in the presence of unknown uncertainties. 
The correlation curve in Figure \ref{fig:bs}\textit{c--d} illustrates that the prediction of the temperature exhibits relatively scattered with a larger deviation. 
This might be because the temperature data, captured at a depth of 12 meters below sea level, remains less susceptible to the influence of other external factors. 
Yet, this real-world dataset considered in this example encompasses numerous elusive variables that are absent from our model's training data which consists solely of velocity and temperature variables. 
Consequently, this limitation hampers accurate predictions of localized high-frequency information, deteriorating the performance of CiGNN. 
Commonly, this issue can be ameliorated by augmenting the training data with pertinent variables to enhance the model's predictive capability. 


\vspace{-3pt}
\subsection{Ablation study}\label{sub:ablation_study}

To gain deeper insights into how each component of the model affects its overall performance, we conducted an ablation study based on the aforementioned datasets.
The results in \textcolor{black}{Table \ref{tab:Ablation Study}} show that the removal of each module leads to  the performance degradation. 
We summarize the observations as follows:
\begin{itemize}
\item Results in Task \#1 demonstrate that, while the promotion may vary, both the conservative and inequality principles play significant roles across different datasets.
\vspace{0pt}
\item Results in Task \#2 indicate that a larger number of edges in the global context leads to more accurate prediction. 
Our proposed trade-off strategy of constructing sparse graphs at different scales greatly improves performance.
\vspace{0pt}
\item Results in Task \#3 show that our temporal learning strategy effectively mitigates error propagation issues with a high-order scheme. In particular, we observe that the performance of the temporal strategy is sensitive to the size of the single-step time interval. 
\end{itemize}

We also consider two variants of CiGNN, namely, CiGNN$^{-}$ and CiGNN$^{*}$, where only the asymmetric information passing is considered in the SymMPNN blocks.
As depicted in \textcolor{black}{Table \ref{tab:performance summary of variants}}, the result shows the efficacy of the symmetric information-passing component. In the case of BS dataset, CiGNN$^{*}$ outperforms CiGNN, which indicates that the effective learning of the inequality term in real-world scenarios is not fully guaranteed. 

Moreover, we provide the ablation results on comparison of various temporal integration strategies shown in Table \ref{tab:temporal strategy}. It is evident that a 4-layer CiGNN model with the simplest 1D convolution filter-based strategy (denoted by C3, with minimal parameter overhead) achieves the best performance, surpassing other strategies including the 2D convolution filtering, TCN and LSTM, demonstrating that our temporal strategy (1D Conv filter) plays a key role in CiGNN to improve the model's long-term rollout prediction accuracy. 




\begin{table}
\caption{
{Results of ablation tests on the temporal integration strategy for the 2D Burgers dataset.}
Here, TCN stands for the temporal convolutional network and LSTM denotes the long short-term memory network. Each model was trained for 1000 iterations with one-step strategy. The integration strategies are denoted by ``P1'' (our proposed 1D convolution filter), ``P2'' (2D convolution filter), ``P3'' (TCN with 1D convolution filter), and ``P4'' (LSTM).}
\label{tab:temporal strategy}
\vspace{-6pt}
\centering
\begin{tabular}{cclcc}
\toprule 
\multirow{2}{*}{\makecell[c]{Case}}  & 
\multirow{2}{*}{\makecell[l]{No. of \\layers} } &
\multirow{2}{*}{\makecell[l]{Temporal \\  strategy} } & 
\multirow{2}{*}{\makecell[l]{No. of \\ parameters}} & 
\multirow{2}{*}{RMSE $\downarrow$}\\
\\
\midrule
C1	&	4     &  None          &514,950       & $9.5812 \times 10^{-2}$   \\
C2	&	16    &  None          &1,901,958     & $1.7593 \times 10^{-1}$   \\
C3	&	4     &  P1 (ours)     &515,007       & $\mathbf{6.9040 \times 10^{-2}}$   \\
C4	&	16    &  P1 (ours)     &1,902,567     & ${1.2442 \times 10^{-1}}$   \\
C5	&	4     &  P2            &514,993       & $8.3591 \times 10^{-2}$   \\
C6	&	16    &  P2            &1,902,001     & $1.6049 \times 10^{-1}$   \\
C7	&	4     &  P3            &514,968       & $8.7441 \times 10^{-2}$   \\
C8	&	16    &  P3            &1,902,012     & $1.6872 \times 10^{-1}$   \\
C9	&	4     &  P4            &779,142       & $1.2820 \times 10^{-1}$   \\
C10	&	16    &  P4            &2,166,150     & $2.2104 \times 10^{-1}$   \\
\bottomrule
\end{tabular}
\vspace{-6pt}
\end{table}


\vspace{-2pt}
\section{Conclusion}\label{Conclusion}
This paper introduces an end-to-end graph-based deep learning model (namely, CiGNN) to predict the evolution of complex spatiotemporal dynamics. 
The CiGNN model is designed to preserve the general differential form of conservation law with a source term based on the fundamental principle of symmetry. 
To this end, we devise a novel SymMPNN module, equipped with symmetric asymmetric information passing in graph edges, for graph representation learning. 
The embedding of such \textit{a prior} knowledge into the network architecture design amounts to adding an inductive bias to the learning process, relaxing the need for heavy training data. 
Moreover, the performance of CiGNN is further lifted by a multi-mesh spatial strategy and a latent temporal integrator. 
The resulting graph learning model is explainable and, meanwhile, the trained model generalizes to predict complex spatiotemporal dynamics over varying ICs, BCs and geometry meshes. The efficacy of the CiGNN model was evaluated in forecasting the dynamics of a variety of spatiotemporal systems (e.g., convection-diffusion system, reaction-diffusion system, flow past a cylinder, and the hydrological dynamics of the Black Sea) based on both synthetic and field-observed datasets. Given limited training data, we demonstrated that the CiGNN model identifies well the underlying dynamics and shows remarkable accuracy. 

Despite its demonstrated efficacy and potential, CiGNN is faced with two main challenges that need to be addressed in the future. On one hand, the brute-force implementation of CiGNN might suffer from the computational bottleneck (e.g., the multi-step rollout training requires heavy memory overhead that depends linearly on the number of rollout steps), which would result in a scalability issue. 
On the other hand, the present model is only deployed to model a classical fluid dynamics problem (e.g., flow past a cylinder), whereas its capacity for predicting highly complex turbulent flows (e.g., large $Re$, complex BCs that induce wall effect, etc.) remains unknown. 
We aim to systematically tackle these challenges in our future study.

\begin{acks}
The work is supported by the National Natural Science Foundation of China (No. 62276269, No. 92270118), the Beijing Natural Science Foundation (No. 1232009), and the Strategic Priority Research Program of the Chinese Academy of Sciences (No. XDB0620103). 
H. Sun and Y. Liu would like to thank the support from the Fundamental Research Funds for the Central Universities (No. 202230265, No. E2EG2202X2). P. Ren would like to disclose that he was involved in this work when he was at Northeastern University, who has not been supported by Huawei Technologies.
\end{acks}

\bibliographystyle{ACM-Reference-Format}
\balance
\bibliography{kdd2025_reference}


\begin{thebibliography}{61}


\ifx \showCODEN    \undefined \def \showCODEN     #1{\unskip}     \fi
\ifx \showDOI      \undefined \def \showDOI       #1{#1}\fi
\ifx \showISBNx    \undefined \def \showISBNx     #1{\unskip}     \fi
\ifx \showISBNxiii \undefined \def \showISBNxiii  #1{\unskip}     \fi
\ifx \showISSN     \undefined \def \showISSN      #1{\unskip}     \fi
\ifx \showLCCN     \undefined \def \showLCCN      #1{\unskip}     \fi
\ifx \shownote     \undefined \def \shownote      #1{#1}          \fi
\ifx \showarticletitle \undefined \def \showarticletitle #1{#1}   \fi
\ifx \showURL      \undefined \def \showURL       {\relax}        \fi
\providecommand\bibfield[2]{#2}
\providecommand\bibinfo[2]{#2}
\providecommand\natexlab[1]{#1}
\providecommand\showeprint[2][]{arXiv:#2}

\bibitem[Alberts and Bilionis(2023)]%
        {alberts2023physics}
\bibfield{author}{\bibinfo{person}{Alex Alberts} {and} \bibinfo{person}{Ilias
  Bilionis}.} \bibinfo{year}{2023}\natexlab{}.
\newblock \showarticletitle{Physics-informed information field theory for
  modeling physical systems with uncertainty quantification}.
\newblock \bibinfo{journal}{\emph{J. Comput. Phys.}}  \bibinfo{volume}{486}
  (\bibinfo{year}{2023}), \bibinfo{pages}{112100}.
\newblock


\bibitem[Ames(2014)]%
        {ames2014numerical}
\bibfield{author}{\bibinfo{person}{William~F Ames}.}
  \bibinfo{year}{2014}\natexlab{}.
\newblock \bibinfo{booktitle}{\emph{Numerical methods for partial differential
  equations}}.
\newblock \bibinfo{publisher}{Academic Press, Inc.}
\newblock


\bibitem[Arzani et~al\mbox{.}(2021)]%
        {arzani2021uncovering}
\bibfield{author}{\bibinfo{person}{Amirhossein Arzani},
  \bibinfo{person}{Jian-Xun Wang}, {and} \bibinfo{person}{Roshan~M D'Souza}.}
  \bibinfo{year}{2021}\natexlab{}.
\newblock \showarticletitle{Uncovering near-wall blood flow from sparse data
  with physics-informed neural networks}.
\newblock \bibinfo{journal}{\emph{Physics of Fluids}} \bibinfo{volume}{33},
  \bibinfo{number}{7} (\bibinfo{year}{2021}).
\newblock


\bibitem[Baggio et~al\mbox{.}(2021)]%
        {baggio2021data}
\bibfield{author}{\bibinfo{person}{Giacomo Baggio}, \bibinfo{person}{Danielle~S
  Bassett}, {and} \bibinfo{person}{Fabio Pasqualetti}.}
  \bibinfo{year}{2021}\natexlab{}.
\newblock \showarticletitle{Data-driven control of complex networks}.
\newblock \bibinfo{journal}{\emph{Nature Communications}} \bibinfo{volume}{12},
  \bibinfo{number}{1} (\bibinfo{year}{2021}), \bibinfo{pages}{1429}.
\newblock


\bibitem[Bi et~al\mbox{.}(2023)]%
        {bi2023accurate}
\bibfield{author}{\bibinfo{person}{Kaifeng Bi}, \bibinfo{person}{Lingxi Xie},
  \bibinfo{person}{Hengheng Zhang}, \bibinfo{person}{Xin Chen},
  \bibinfo{person}{Xiaotao Gu}, {and} \bibinfo{person}{Qi Tian}.}
  \bibinfo{year}{2023}\natexlab{}.
\newblock \showarticletitle{Accurate medium-range global weather forecasting
  with 3D neural networks}.
\newblock \bibinfo{journal}{\emph{Nature}} \bibinfo{volume}{619},
  \bibinfo{number}{7970} (\bibinfo{year}{2023}), \bibinfo{pages}{533--538}.
\newblock


\bibitem[Boull{\'e} et~al\mbox{.}(2023)]%
        {boulle2023elliptic}
\bibfield{author}{\bibinfo{person}{Nicolas Boull{\'e}}, \bibinfo{person}{Diana
  Halikias}, {and} \bibinfo{person}{Alex Townsend}.}
  \bibinfo{year}{2023}\natexlab{}.
\newblock \showarticletitle{Elliptic PDE learning is provably data-efficient}.
\newblock \bibinfo{journal}{\emph{Proceedings of the National Academy of
  Sciences}} \bibinfo{volume}{120}, \bibinfo{number}{39}
  (\bibinfo{year}{2023}), \bibinfo{pages}{e2303904120}.
\newblock


\bibitem[Brandstetter et~al\mbox{.}(2022)]%
        {brandstetter2022message}
\bibfield{author}{\bibinfo{person}{Johannes Brandstetter},
  \bibinfo{person}{Daniel Worrall}, {and} \bibinfo{person}{Max Welling}.}
  \bibinfo{year}{2022}\natexlab{}.
\newblock \showarticletitle{Message passing neural PDE solvers}. In
  \bibinfo{booktitle}{\emph{International Conference on Learning
  Representations}}.
\newblock


\bibitem[Brunton et~al\mbox{.}(2020)]%
        {brunton2020machine}
\bibfield{author}{\bibinfo{person}{Steven~L Brunton}, \bibinfo{person}{Bernd~R
  Noack}, {and} \bibinfo{person}{Petros Koumoutsakos}.}
  \bibinfo{year}{2020}\natexlab{}.
\newblock \showarticletitle{Machine learning for fluid mechanics}.
\newblock \bibinfo{journal}{\emph{Annual Review of Fluid Mechanics}}
  \bibinfo{volume}{52} (\bibinfo{year}{2020}), \bibinfo{pages}{477--508}.
\newblock


\bibitem[Brunton et~al\mbox{.}(2016)]%
        {brunton2016discovering}
\bibfield{author}{\bibinfo{person}{Steven~L Brunton}, \bibinfo{person}{Joshua~L
  Proctor}, {and} \bibinfo{person}{J~Nathan Kutz}.}
  \bibinfo{year}{2016}\natexlab{}.
\newblock \showarticletitle{Discovering governing equations from data by sparse
  identification of nonlinear dynamical systems}.
\newblock \bibinfo{journal}{\emph{Proceedings of the National Academy of
  Sciences}} \bibinfo{volume}{113}, \bibinfo{number}{15}
  (\bibinfo{year}{2016}), \bibinfo{pages}{3932--3937}.
\newblock


\bibitem[Chen et~al\mbox{.}(2021)]%
        {chen2021physics}
\bibfield{author}{\bibinfo{person}{Zhao Chen}, \bibinfo{person}{Yang Liu},
  {and} \bibinfo{person}{Hao Sun}.} \bibinfo{year}{2021}\natexlab{}.
\newblock \showarticletitle{Physics-informed learning of governing equations
  from scarce data}.
\newblock \bibinfo{journal}{\emph{Nature Communications}} \bibinfo{volume}{12},
  \bibinfo{number}{1} (\bibinfo{year}{2021}), \bibinfo{pages}{6136}.
\newblock


\bibitem[Eymard et~al\mbox{.}(2000)]%
        {eymard2000finite}
\bibfield{author}{\bibinfo{person}{Robert Eymard}, \bibinfo{person}{Thierry
  Gallou{\"e}t}, {and} \bibinfo{person}{Rapha{\`e}le Herbin}.}
  \bibinfo{year}{2000}\natexlab{}.
\newblock \showarticletitle{Finite volume methods}.
\newblock \bibinfo{journal}{\emph{Handbook of numerical analysis}}
  \bibinfo{volume}{7} (\bibinfo{year}{2000}), \bibinfo{pages}{713--1018}.
\newblock


\bibitem[Farina et~al\mbox{.}(2015)]%
        {FARINA2015631}
\bibfield{author}{\bibinfo{person}{R. Farina}, \bibinfo{person}{S. Dobricic},
  \bibinfo{person}{A. Storto}, \bibinfo{person}{S. Masina}, {and}
  \bibinfo{person}{S. Cuomo}.} \bibinfo{year}{2015}\natexlab{}.
\newblock \showarticletitle{A revised scheme to compute horizontal covariances
  in an oceanographic 3D-VAR assimilation system}.
\newblock \bibinfo{journal}{\emph{J. Comput. Phys.}}  \bibinfo{volume}{284}
  (\bibinfo{year}{2015}), \bibinfo{pages}{631--647}.
\newblock
\showISSN{0021-9991}
\urldef\tempurl%
\url{https://doi.org/10.1016/j.jcp.2015.01.003}
\showDOI{\tempurl}


\bibitem[Floryan and Graham(2022)]%
        {floryan2022data}
\bibfield{author}{\bibinfo{person}{Daniel Floryan} {and}
  \bibinfo{person}{Michael~D Graham}.} \bibinfo{year}{2022}\natexlab{}.
\newblock \showarticletitle{Data-driven discovery of intrinsic dynamics}.
\newblock \bibinfo{journal}{\emph{Nature Machine Intelligence}}
  \bibinfo{volume}{4}, \bibinfo{number}{12} (\bibinfo{year}{2022}),
  \bibinfo{pages}{1113--1120}.
\newblock


\bibitem[Grega et~al\mbox{.}(2024)]%
        {grega2024energy}
\bibfield{author}{\bibinfo{person}{Ivan Grega}, \bibinfo{person}{Ilyes
  Batatia}, \bibinfo{person}{G{\'a}bor Cs{\'a}nyi}, \bibinfo{person}{Sri
  Karlapati}, {and} \bibinfo{person}{Vikram~S Deshpande}.}
  \bibinfo{year}{2024}\natexlab{}.
\newblock \showarticletitle{Energy-conserving equivariant GNN for elasticity of
  lattice architected metamaterials}.
\newblock \bibinfo{journal}{\emph{arXiv preprint arXiv:2401.16914}}
  (\bibinfo{year}{2024}).
\newblock


\bibitem[Han et~al\mbox{.}(2022)]%
        {han2022learning}
\bibfield{author}{\bibinfo{person}{Jiaqi Han}, \bibinfo{person}{Wenbing Huang},
  \bibinfo{person}{Hengbo Ma}, \bibinfo{person}{Jiachen Li},
  \bibinfo{person}{Josh Tenenbaum}, {and} \bibinfo{person}{Chuang Gan}.}
  \bibinfo{year}{2022}\natexlab{}.
\newblock \showarticletitle{Learning physical dynamics with subequivariant
  graph neural networks}.
\newblock \bibinfo{journal}{\emph{Advances in Neural Information Processing
  Systems}}  \bibinfo{volume}{35} (\bibinfo{year}{2022}),
  \bibinfo{pages}{26256--26268}.
\newblock


\bibitem[Hao et~al\mbox{.}(2023)]%
        {hao2023gnot}
\bibfield{author}{\bibinfo{person}{Zhongkai Hao}, \bibinfo{person}{Zhengyi
  Wang}, \bibinfo{person}{Hang Su}, \bibinfo{person}{Chengyang Ying},
  \bibinfo{person}{Yinpeng Dong}, \bibinfo{person}{Songming Liu},
  \bibinfo{person}{Ze Cheng}, \bibinfo{person}{Jian Song}, {and}
  \bibinfo{person}{Jun Zhu}.} \bibinfo{year}{2023}\natexlab{}.
\newblock \showarticletitle{Gnot: A general neural operator transformer for
  operator learning}. In \bibinfo{booktitle}{\emph{International Conference on
  Machine Learning}}. PMLR, \bibinfo{pages}{12556--12569}.
\newblock


\bibitem[Hernández et~al\mbox{.}(2023)]%
        {hernandez2023thermodynamics}
\bibfield{author}{\bibinfo{person}{Quercus Hernández},
  \bibinfo{person}{Alberto Badías}, \bibinfo{person}{Francisco Chinesta},
  {and} \bibinfo{person}{Elías Cueto}.} \bibinfo{year}{2023}\natexlab{}.
\newblock \showarticletitle{Thermodynamics-informed neural networks for
  physically realistic mixed reality}.
\newblock \bibinfo{journal}{\emph{Computer Methods in Applied Mechanics and
  Engineering}}  \bibinfo{volume}{407} (\bibinfo{year}{2023}),
  \bibinfo{pages}{115912}.
\newblock
\showISSN{0045-7825}
\urldef\tempurl%
\url{https://doi.org/10.1016/j.cma.2023.115912}
\showDOI{\tempurl}


\bibitem[Hersbach et~al\mbox{.}(2020)]%
        {hersbach2020era5}
\bibfield{author}{\bibinfo{person}{Hans Hersbach}, \bibinfo{person}{Bill Bell},
  \bibinfo{person}{Paul Berrisford}, \bibinfo{person}{Shoji Hirahara},
  \bibinfo{person}{Andr{\'a}s Hor{\'a}nyi}, \bibinfo{person}{Joaqu{\'\i}n
  Mu{\~n}oz-Sabater}, \bibinfo{person}{Julien Nicolas}, \bibinfo{person}{Carole
  Peubey}, \bibinfo{person}{Raluca Radu}, \bibinfo{person}{Dinand Schepers},
  {et~al\mbox{.}}} \bibinfo{year}{2020}\natexlab{}.
\newblock \showarticletitle{The ERA5 global reanalysis}.
\newblock \bibinfo{journal}{\emph{Quarterly Journal of the Royal Meteorological
  Society}} \bibinfo{volume}{146}, \bibinfo{number}{730}
  (\bibinfo{year}{2020}), \bibinfo{pages}{1999--2049}.
\newblock


\bibitem[Hirani(2003)]%
        {hirani2003discrete}
\bibfield{author}{\bibinfo{person}{Anil~Nirmal Hirani}.}
  \bibinfo{year}{2003}\natexlab{}.
\newblock \bibinfo{booktitle}{\emph{Discrete exterior calculus}}.
\newblock \bibinfo{publisher}{California Institute of Technology}.
\newblock


\bibitem[Horie and Mitsume(2024)]%
        {horie2024graph}
\bibfield{author}{\bibinfo{person}{Masanobu Horie} {and} \bibinfo{person}{Naoto
  Mitsume}.} \bibinfo{year}{2024}\natexlab{}.
\newblock \showarticletitle{Graph Neural PDE Solvers with Conservation and
  Similarity-Equivariance}.
\newblock \bibinfo{journal}{\emph{arXiv preprint arXiv:2405.16183}}
  (\bibinfo{year}{2024}).
\newblock


\bibitem[Jagtap et~al\mbox{.}(2020)]%
        {jagtap2020conservative}
\bibfield{author}{\bibinfo{person}{Ameya~D Jagtap}, \bibinfo{person}{Ehsan
  Kharazmi}, {and} \bibinfo{person}{George~Em Karniadakis}.}
  \bibinfo{year}{2020}\natexlab{}.
\newblock \showarticletitle{Conservative physics-informed neural networks on
  discrete domains for conservation laws: Applications to forward and inverse
  problems}.
\newblock \bibinfo{journal}{\emph{Computer Methods in Applied Mechanics and
  Engineering}}  \bibinfo{volume}{365} (\bibinfo{year}{2020}),
  \bibinfo{pages}{113028}.
\newblock


\bibitem[Karniadakis et~al\mbox{.}(2021)]%
        {karniadakis2021physics}
\bibfield{author}{\bibinfo{person}{George~Em Karniadakis},
  \bibinfo{person}{Ioannis~G Kevrekidis}, \bibinfo{person}{Lu Lu},
  \bibinfo{person}{Paris Perdikaris}, \bibinfo{person}{Sifan Wang}, {and}
  \bibinfo{person}{Liu Yang}.} \bibinfo{year}{2021}\natexlab{}.
\newblock \showarticletitle{Physics-informed machine learning}.
\newblock \bibinfo{journal}{\emph{Nature Reviews Physics}} \bibinfo{volume}{3},
  \bibinfo{number}{6} (\bibinfo{year}{2021}), \bibinfo{pages}{422--440}.
\newblock


\bibitem[Kingma and Ba(2015)]%
        {kingmaB2015adam}
\bibfield{author}{\bibinfo{person}{Diederik~P. Kingma} {and}
  \bibinfo{person}{Jimmy Ba}.} \bibinfo{year}{2015}\natexlab{}.
\newblock \showarticletitle{Adam: {A} Method for Stochastic Optimization}. In
  \bibinfo{booktitle}{\emph{International Conference on Learning
  Representations}}.
\newblock


\bibitem[Kochkov et~al\mbox{.}(2021)]%
        {kochkov2021machine}
\bibfield{author}{\bibinfo{person}{Dmitrii Kochkov}, \bibinfo{person}{Jamie~A
  Smith}, \bibinfo{person}{Ayya Alieva}, \bibinfo{person}{Qing Wang},
  \bibinfo{person}{Michael~P Brenner}, {and} \bibinfo{person}{Stephan Hoyer}.}
  \bibinfo{year}{2021}\natexlab{}.
\newblock \showarticletitle{Machine learning--accelerated computational fluid
  dynamics}.
\newblock \bibinfo{journal}{\emph{Proceedings of the National Academy of
  Sciences}} \bibinfo{volume}{118}, \bibinfo{number}{21}
  (\bibinfo{year}{2021}), \bibinfo{pages}{e2101784118}.
\newblock


\bibitem[Lam et~al\mbox{.}(2023)]%
        {lam2023learning}
\bibfield{author}{\bibinfo{person}{Remi Lam}, \bibinfo{person}{Alvaro
  Sanchez-Gonzalez}, \bibinfo{person}{Matthew Willson}, \bibinfo{person}{Peter
  Wirnsberger}, \bibinfo{person}{Meire Fortunato}, \bibinfo{person}{Ferran
  Alet}, \bibinfo{person}{Suman Ravuri}, \bibinfo{person}{Timo Ewalds},
  \bibinfo{person}{Zach Eaton-Rosen}, \bibinfo{person}{Weihua Hu},
  {et~al\mbox{.}}} \bibinfo{year}{2023}\natexlab{}.
\newblock \showarticletitle{Learning skillful medium-range global weather
  forecasting}.
\newblock \bibinfo{journal}{\emph{Science}} \bibinfo{volume}{382},
  \bibinfo{number}{6677} (\bibinfo{year}{2023}), \bibinfo{pages}{1416--1421}.
\newblock


\bibitem[Lanthaler et~al\mbox{.}(2023)]%
        {lanthaler2023nonlinear}
\bibfield{author}{\bibinfo{person}{Samuel Lanthaler}, \bibinfo{person}{Roberto
  Molinaro}, \bibinfo{person}{Patrik Hadorn}, {and} \bibinfo{person}{Siddhartha
  Mishra}.} \bibinfo{year}{2023}\natexlab{}.
\newblock \showarticletitle{Nonlinear Reconstruction for Operator Learning of
  PDEs with Discontinuities}. In \bibinfo{booktitle}{\emph{International
  Conference on Learning Representations}}.
\newblock


\bibitem[Li et~al\mbox{.}(2021)]%
        {li2021Fourier}
\bibfield{author}{\bibinfo{person}{Z. Li}, \bibinfo{person}{N.~B. Kovachki},
  \bibinfo{person}{K. Azizzadenesheli}, \bibinfo{person}{B. Liu},
  \bibinfo{person}{K. Bhattacharya}, \bibinfo{person}{A. Stuart}, {and}
  \bibinfo{person}{A. Anandkumar}.} \bibinfo{year}{2021}\natexlab{}.
\newblock \showarticletitle{Fourier Neural Operator for Parametric Partial
  Differential Equations}. In \bibinfo{booktitle}{\emph{International
  Conference on Learning Representations}}.
\newblock


\bibitem[Liu et~al\mbox{.}(2024)]%
        {liu2024segno}
\bibfield{author}{\bibinfo{person}{Yang Liu}, \bibinfo{person}{Jiashun Cheng},
  \bibinfo{person}{Haihong Zhao}, \bibinfo{person}{Tingyang Xu},
  \bibinfo{person}{Peilin Zhao}, \bibinfo{person}{Fugee Tsung},
  \bibinfo{person}{Jia Li}, {and} \bibinfo{person}{Yu Rong}.}
  \bibinfo{year}{2024}\natexlab{}.
\newblock \showarticletitle{SEGNO: Generalizing Equivariant Graph Neural
  Networks with Physical Inductive Biases}. In \bibinfo{booktitle}{\emph{The
  Twelfth International Conference on Learning Representations}}.
\newblock


\bibitem[Lu et~al\mbox{.}(2021a)]%
        {lu2021learning}
\bibfield{author}{\bibinfo{person}{Lu Lu}, \bibinfo{person}{Pengzhan Jin},
  \bibinfo{person}{Guofei Pang}, \bibinfo{person}{Zhongqiang Zhang}, {and}
  \bibinfo{person}{George~Em Karniadakis}.} \bibinfo{year}{2021}\natexlab{a}.
\newblock \showarticletitle{Learning nonlinear operators via DeepONet based on
  the universal approximation theorem of operators}.
\newblock \bibinfo{journal}{\emph{Nature Machine Intelligence}}
  \bibinfo{volume}{3}, \bibinfo{number}{3} (\bibinfo{year}{2021}),
  \bibinfo{pages}{218--229}.
\newblock


\bibitem[Lu et~al\mbox{.}(2021b)]%
        {lu2019deeponet}
\bibfield{author}{\bibinfo{person}{Lu Lu}, \bibinfo{person}{Pengzhan Jin},
  \bibinfo{person}{Guofei Pang}, \bibinfo{person}{Zhongqiang Zhang}, {and}
  \bibinfo{person}{George~Em Karniadakis}.} \bibinfo{year}{2021}\natexlab{b}.
\newblock \showarticletitle{Learning nonlinear operators via {DeepONet} based
  on the universal approximation theorem of operators}.
\newblock \bibinfo{journal}{\emph{Nature Machine Intelligence}}
  \bibinfo{volume}{3}, \bibinfo{number}{3} (\bibinfo{year}{2021}),
  \bibinfo{pages}{218--229}.
\newblock
\urldef\tempurl%
\url{https://doi.org/10.1038/s42256-021-00302-5}
\showDOI{\tempurl}


\bibitem[Niu et~al\mbox{.}(2023)]%
        {niu2023modeling}
\bibfield{author}{\bibinfo{person}{Sijun Niu}, \bibinfo{person}{Enrui Zhang},
  \bibinfo{person}{Yuri Bazilevs}, {and} \bibinfo{person}{Vikas Srivastava}.}
  \bibinfo{year}{2023}\natexlab{}.
\newblock \showarticletitle{Modeling finite-strain plasticity using
  physics-informed neural network and assessment of the network performance}.
\newblock \bibinfo{journal}{\emph{Journal of the Mechanics and Physics of
  Solids}}  \bibinfo{volume}{172} (\bibinfo{year}{2023}),
  \bibinfo{pages}{105177}.
\newblock


\bibitem[Pan et~al\mbox{.}(2023)]%
        {pan2023neural}
\bibfield{author}{\bibinfo{person}{Shaowu Pan}, \bibinfo{person}{Steven~L
  Brunton}, {and} \bibinfo{person}{J~Nathan Kutz}.}
  \bibinfo{year}{2023}\natexlab{}.
\newblock \showarticletitle{Neural implicit flow: a mesh-agnostic
  dimensionality reduction paradigm of spatio-temporal data}.
\newblock \bibinfo{journal}{\emph{Journal of Machine Learning Research}}
  \bibinfo{volume}{24}, \bibinfo{number}{41} (\bibinfo{year}{2023}),
  \bibinfo{pages}{1--60}.
\newblock


\bibitem[Paszke et~al\mbox{.}(2019)]%
        {Paszke2019PyTorch}
\bibfield{author}{\bibinfo{person}{Adam Paszke}, \bibinfo{person}{Sam Gross},
  \bibinfo{person}{Francisco Massa}, \bibinfo{person}{Adam Lerer},
  \bibinfo{person}{James Bradbury}, \bibinfo{person}{Gregory Chanan},
  \bibinfo{person}{Trevor Killeen}, \bibinfo{person}{Zeming Lin},
  \bibinfo{person}{Natalia Gimelshein}, \bibinfo{person}{Luca Antiga},
  {et~al\mbox{.}}} \bibinfo{year}{2019}\natexlab{}.
\newblock \showarticletitle{Pytorch: An imperative style, high-performance deep
  learning library}.
\newblock \bibinfo{journal}{\emph{Advances in neural information processing
  systems}}  \bibinfo{volume}{32} (\bibinfo{year}{2019}).
\newblock


\bibitem[Pfaff et~al\mbox{.}(2021)]%
        {pfaff2021Learning}
\bibfield{author}{\bibinfo{person}{T. Pfaff}, \bibinfo{person}{M. Fortunato},
  \bibinfo{person}{A. Sanchez-Gonzalez}, {and} \bibinfo{person}{P. Battaglia}.}
  \bibinfo{year}{2021}\natexlab{}.
\newblock \showarticletitle{Learning Mesh-Based Simulation with Graph
  Networks}. In \bibinfo{booktitle}{\emph{International Conference on Learning
  Representations}}.
\newblock


\bibitem[Pineda et~al\mbox{.}(2023)]%
        {pineda2023geometric}
\bibfield{author}{\bibinfo{person}{Jes{\'u}s Pineda}, \bibinfo{person}{Benjamin
  Midtvedt}, \bibinfo{person}{Harshith Bachimanchi}, \bibinfo{person}{Sergio
  No{\'e}}, \bibinfo{person}{Daniel Midtvedt}, \bibinfo{person}{Giovanni
  Volpe}, {and} \bibinfo{person}{Carlo Manzo}.}
  \bibinfo{year}{2023}\natexlab{}.
\newblock \showarticletitle{Geometric deep learning reveals the spatiotemporal
  features of microscopic motion}.
\newblock \bibinfo{journal}{\emph{Nature Machine Intelligence}}
  \bibinfo{volume}{5}, \bibinfo{number}{1} (\bibinfo{year}{2023}),
  \bibinfo{pages}{71--82}.
\newblock


\bibitem[Raissi et~al\mbox{.}(2019)]%
        {raissi2019physics}
\bibfield{author}{\bibinfo{person}{Maziar Raissi}, \bibinfo{person}{Paris
  Perdikaris}, {and} \bibinfo{person}{George~E Karniadakis}.}
  \bibinfo{year}{2019}\natexlab{}.
\newblock \showarticletitle{Physics-informed neural networks: A deep learning
  framework for solving forward and inverse problems involving nonlinear
  partial differential equations}.
\newblock \bibinfo{journal}{\emph{Journal of Computational physics}}
  \bibinfo{volume}{378} (\bibinfo{year}{2019}), \bibinfo{pages}{686--707}.
\newblock


\bibitem[Raissi et~al\mbox{.}(2020)]%
        {raissi2020hidden}
\bibfield{author}{\bibinfo{person}{Maziar Raissi}, \bibinfo{person}{Alireza
  Yazdani}, {and} \bibinfo{person}{George~Em Karniadakis}.}
  \bibinfo{year}{2020}\natexlab{}.
\newblock \showarticletitle{Hidden fluid mechanics: Learning velocity and
  pressure fields from flow visualizations}.
\newblock \bibinfo{journal}{\emph{Science}} \bibinfo{volume}{367},
  \bibinfo{number}{6481} (\bibinfo{year}{2020}), \bibinfo{pages}{1026--1030}.
\newblock


\bibitem[Rao et~al\mbox{.}(2023)]%
        {rao2023encoding}
\bibfield{author}{\bibinfo{person}{Chengping Rao}, \bibinfo{person}{Pu Ren},
  \bibinfo{person}{Qi Wang}, \bibinfo{person}{Oral Buyukozturk},
  \bibinfo{person}{Hao Sun}, {and} \bibinfo{person}{Yang Liu}.}
  \bibinfo{year}{2023}\natexlab{}.
\newblock \showarticletitle{Encoding physics to learn reaction--diffusion
  processes}.
\newblock \bibinfo{journal}{\emph{Nature Machine Intelligence}}
  \bibinfo{volume}{5}, \bibinfo{number}{7} (\bibinfo{year}{2023}),
  \bibinfo{pages}{765--779}.
\newblock


\bibitem[Rao et~al\mbox{.}(2021)]%
        {rao2021physics}
\bibfield{author}{\bibinfo{person}{Chengping Rao}, \bibinfo{person}{Hao Sun},
  {and} \bibinfo{person}{Yang Liu}.} \bibinfo{year}{2021}\natexlab{}.
\newblock \showarticletitle{Physics-informed deep learning for computational
  elastodynamics without labeled data}.
\newblock \bibinfo{journal}{\emph{Journal of Engineering Mechanics}}
  \bibinfo{volume}{147}, \bibinfo{number}{8} (\bibinfo{year}{2021}),
  \bibinfo{pages}{04021043}.
\newblock


\bibitem[Ravuri et~al\mbox{.}(2021)]%
        {ravuri2021skilful}
\bibfield{author}{\bibinfo{person}{Suman Ravuri}, \bibinfo{person}{Karel Lenc},
  \bibinfo{person}{Matthew Willson}, \bibinfo{person}{Dmitry Kangin},
  \bibinfo{person}{Remi Lam}, \bibinfo{person}{Piotr Mirowski},
  \bibinfo{person}{Megan Fitzsimons}, \bibinfo{person}{Maria Athanassiadou},
  \bibinfo{person}{Sheleem Kashem}, \bibinfo{person}{Sam Madge},
  {et~al\mbox{.}}} \bibinfo{year}{2021}\natexlab{}.
\newblock \showarticletitle{Skilful precipitation nowcasting using deep
  generative models of radar}.
\newblock \bibinfo{journal}{\emph{Nature}} \bibinfo{volume}{597},
  \bibinfo{number}{7878} (\bibinfo{year}{2021}), \bibinfo{pages}{672--677}.
\newblock


\bibitem[Regazzoni et~al\mbox{.}(2019)]%
        {regazzoni2019machine}
\bibfield{author}{\bibinfo{person}{Francesco Regazzoni}, \bibinfo{person}{Luca
  Dede}, {and} \bibinfo{person}{Alfio Quarteroni}.}
  \bibinfo{year}{2019}\natexlab{}.
\newblock \showarticletitle{Machine learning for fast and reliable solution of
  time-dependent differential equations}.
\newblock \bibinfo{journal}{\emph{Journal of Computational physics}}
  \bibinfo{volume}{397} (\bibinfo{year}{2019}), \bibinfo{pages}{108852}.
\newblock


\bibitem[Ren et~al\mbox{.}(2024)]%
        {ren2024seismicnet}
\bibfield{author}{\bibinfo{person}{Pu Ren}, \bibinfo{person}{Chengping Rao},
  \bibinfo{person}{Su Chen}, \bibinfo{person}{Jian-Xun Wang},
  \bibinfo{person}{Hao Sun}, {and} \bibinfo{person}{Yang Liu}.}
  \bibinfo{year}{2024}\natexlab{}.
\newblock \showarticletitle{SeismicNet: Physics-informed neural networks for
  seismic wave modeling in semi-infinite domain}.
\newblock \bibinfo{journal}{\emph{Computer Physics Communications}}
  \bibinfo{volume}{295} (\bibinfo{year}{2024}), \bibinfo{pages}{109010}.
\newblock


\bibitem[Ren et~al\mbox{.}(2023)]%
        {ren2023physr}
\bibfield{author}{\bibinfo{person}{Pu Ren}, \bibinfo{person}{Chengping Rao},
  \bibinfo{person}{Yang Liu}, \bibinfo{person}{Zihan Ma}, \bibinfo{person}{Qi
  Wang}, \bibinfo{person}{Jian-Xun Wang}, {and} \bibinfo{person}{Hao Sun}.}
  \bibinfo{year}{2023}\natexlab{}.
\newblock \showarticletitle{{PhySR}: Physics-informed deep super-resolution for
  spatiotemporal data}.
\newblock \bibinfo{journal}{\emph{J. Comput. Phys.}}  \bibinfo{volume}{492}
  (\bibinfo{year}{2023}), \bibinfo{pages}{112438}.
\newblock


\bibitem[Ren et~al\mbox{.}(2022)]%
        {ren2022phycrnet}
\bibfield{author}{\bibinfo{person}{Pu Ren}, \bibinfo{person}{Chengping Rao},
  \bibinfo{person}{Yang Liu}, \bibinfo{person}{Jian-Xun Wang}, {and}
  \bibinfo{person}{Hao Sun}.} \bibinfo{year}{2022}\natexlab{}.
\newblock \showarticletitle{PhyCRNet: Physics-informed convolutional-recurrent
  network for solving spatiotemporal PDEs}.
\newblock \bibinfo{journal}{\emph{Computer Methods in Applied Mechanics and
  Engineering}}  \bibinfo{volume}{389} (\bibinfo{year}{2022}),
  \bibinfo{pages}{114399}.
\newblock


\bibitem[Rezaei et~al\mbox{.}(2024)]%
        {rezaei2024learning}
\bibfield{author}{\bibinfo{person}{Shahed Rezaei}, \bibinfo{person}{Ahmad
  Moeineddin}, {and} \bibinfo{person}{Ali Harandi}.}
  \bibinfo{year}{2024}\natexlab{}.
\newblock \showarticletitle{Learning solutions of thermodynamics-based
  nonlinear constitutive material models using physics-informed neural
  networks}.
\newblock \bibinfo{journal}{\emph{Computational Mechanics}}
  (\bibinfo{year}{2024}), \bibinfo{pages}{1--34}.
\newblock


\bibitem[Sanchez-Gonzalez et~al\mbox{.}(2020)]%
        {sanchez2020learning}
\bibfield{author}{\bibinfo{person}{Alvaro Sanchez-Gonzalez},
  \bibinfo{person}{Jonathan Godwin}, \bibinfo{person}{Tobias Pfaff},
  \bibinfo{person}{Rex Ying}, \bibinfo{person}{Jure Leskovec}, {and}
  \bibinfo{person}{Peter Battaglia}.} \bibinfo{year}{2020}\natexlab{}.
\newblock \showarticletitle{Learning to simulate complex physics with graph
  networks}. In \bibinfo{booktitle}{\emph{International Conference on Machine
  Learning}}. \bibinfo{pages}{8459--8468}.
\newblock


\bibitem[Sarlet and Cantrijn(1981)]%
        {sarlet1981generalizations}
\bibfield{author}{\bibinfo{person}{Willy Sarlet} {and} \bibinfo{person}{Frans
  Cantrijn}.} \bibinfo{year}{1981}\natexlab{}.
\newblock \showarticletitle{Generalizations of {Noether’s} theorem in
  classical mechanics}.
\newblock \bibinfo{journal}{\emph{Siam Review}} \bibinfo{volume}{23},
  \bibinfo{number}{4} (\bibinfo{year}{1981}), \bibinfo{pages}{467--494}.
\newblock


\bibitem[Schmidt and Lipson(2009)]%
        {schmidt2009distilling}
\bibfield{author}{\bibinfo{person}{Michael Schmidt} {and} \bibinfo{person}{Hod
  Lipson}.} \bibinfo{year}{2009}\natexlab{}.
\newblock \showarticletitle{Distilling free-form natural laws from experimental
  data}.
\newblock \bibinfo{journal}{\emph{Science}} \bibinfo{volume}{324},
  \bibinfo{number}{5923} (\bibinfo{year}{2009}), \bibinfo{pages}{81--85}.
\newblock


\bibitem[Tadmor(1984)]%
        {tadmor1984skew}
\bibfield{author}{\bibinfo{person}{Eitan Tadmor}.}
  \bibinfo{year}{1984}\natexlab{}.
\newblock \showarticletitle{Skew-selfadjoint form for systems of conservation
  laws}.
\newblock \bibinfo{journal}{\emph{J. Math. Anal. Appl.}} \bibinfo{volume}{103},
  \bibinfo{number}{2} (\bibinfo{year}{1984}), \bibinfo{pages}{428--442}.
\newblock


\bibitem[Tran et~al\mbox{.}(2023)]%
        {tran2023factorized}
\bibfield{author}{\bibinfo{person}{Alasdair Tran}, \bibinfo{person}{Alexander
  Mathews}, \bibinfo{person}{Lexing Xie}, {and} \bibinfo{person}{Cheng~Soon
  Ong}.} \bibinfo{year}{2023}\natexlab{}.
\newblock \showarticletitle{Factorized {Fourier} Neural Operators}. In
  \bibinfo{booktitle}{\emph{International Conference on Learning
  Representations}}.
\newblock


\bibitem[Wang et~al\mbox{.}(2024a)]%
        {wang2024beno}
\bibfield{author}{\bibinfo{person}{Haixin Wang}, \bibinfo{person}{LI Jiaxin},
  \bibinfo{person}{Anubhav Dwivedi}, \bibinfo{person}{Kentaro Hara}, {and}
  \bibinfo{person}{Tailin Wu}.} \bibinfo{year}{2024}\natexlab{a}.
\newblock \showarticletitle{{BENO}: Boundary-embedded Neural Operators for
  Elliptic PDEs}. In \bibinfo{booktitle}{\emph{The Twelfth International
  Conference on Learning Representations}}.
\newblock


\bibitem[Wang et~al\mbox{.}(2024b)]%
        {wangp2024}
\bibfield{author}{\bibinfo{person}{Qi Wang}, \bibinfo{person}{Pu Ren},
  \bibinfo{person}{Hao Zhou}, \bibinfo{person}{Xin-Yang Liu},
  \bibinfo{person}{Zhiwen Deng}, \bibinfo{person}{Yi Zhang},
  \bibinfo{person}{Ruizhi Chengze}, \bibinfo{person}{Hongsheng Liu},
  \bibinfo{person}{Zidong Wang}, \bibinfo{person}{Jian-Xun Wang},
  {et~al\mbox{.}}} \bibinfo{year}{2024}\natexlab{b}.
\newblock \showarticletitle{P$^2$C$^2$ Net: PDE-Preserved Coarse Correction
  Network for efficient prediction of spatiotemporal dynamics}. In
  \bibinfo{booktitle}{\emph{Advances in Neural Information Processing
  Systems}}.
\newblock


\bibitem[Wu et~al\mbox{.}(2024a)]%
        {wu2024equivariant}
\bibfield{author}{\bibinfo{person}{Liming Wu}, \bibinfo{person}{Zhichao Hou},
  \bibinfo{person}{Jirui Yuan}, \bibinfo{person}{Yu Rong}, {and}
  \bibinfo{person}{Wenbing Huang}.} \bibinfo{year}{2024}\natexlab{a}.
\newblock \showarticletitle{Equivariant spatio-temporal attentive graph
  networks to simulate physical dynamics}.
\newblock \bibinfo{journal}{\emph{Advances in Neural Information Processing
  Systems}}  \bibinfo{volume}{36} (\bibinfo{year}{2024}).
\newblock


\bibitem[Wu et~al\mbox{.}(2022)]%
        {wu2022learning}
\bibfield{author}{\bibinfo{person}{Tailin Wu}, \bibinfo{person}{Takashi
  Maruyama}, {and} \bibinfo{person}{Jure Leskovec}.}
  \bibinfo{year}{2022}\natexlab{}.
\newblock \showarticletitle{Learning to accelerate partial differential
  equations via latent global evolution}.
\newblock \bibinfo{journal}{\emph{Advances in Neural Information Processing
  Systems}}  \bibinfo{volume}{35} (\bibinfo{year}{2022}),
  \bibinfo{pages}{2240--2253}.
\newblock


\bibitem[Wu et~al\mbox{.}(2024b)]%
        {wu2024uncertainty}
\bibfield{author}{\bibinfo{person}{Tailin Wu}, \bibinfo{person}{Willie
  Neiswanger}, \bibinfo{person}{Hongtao Zheng}, \bibinfo{person}{Stefano
  Ermon}, {and} \bibinfo{person}{Jure Leskovec}.}
  \bibinfo{year}{2024}\natexlab{b}.
\newblock \showarticletitle{Uncertainty Quantification for Forward and Inverse
  Problems of PDEs via Latent Global Evolution}. In
  \bibinfo{booktitle}{\emph{Proceedings of the AAAI Conference on Artificial
  Intelligence}}, Vol.~\bibinfo{volume}{38}. \bibinfo{pages}{320--328}.
\newblock


\bibitem[Wu et~al\mbox{.}(2020)]%
        {wu2020comprehensive}
\bibfield{author}{\bibinfo{person}{Zonghan Wu}, \bibinfo{person}{Shirui Pan},
  \bibinfo{person}{Fengwen Chen}, \bibinfo{person}{Guodong Long},
  \bibinfo{person}{Chengqi Zhang}, {and} \bibinfo{person}{S~Yu Philip}.}
  \bibinfo{year}{2020}\natexlab{}.
\newblock \showarticletitle{A comprehensive survey on graph neural networks}.
\newblock \bibinfo{journal}{\emph{IEEE transactions on neural networks and
  learning systems}} \bibinfo{volume}{32}, \bibinfo{number}{1}
  (\bibinfo{year}{2020}), \bibinfo{pages}{4--24}.
\newblock


\bibitem[Zhai et~al\mbox{.}(2023)]%
        {zhai2023model}
\bibfield{author}{\bibinfo{person}{Zheng-Meng Zhai},
  \bibinfo{person}{Mohammadamin Moradi}, \bibinfo{person}{Ling-Wei Kong},
  \bibinfo{person}{Bryan Glaz}, \bibinfo{person}{Mulugeta Haile}, {and}
  \bibinfo{person}{Ying-Cheng Lai}.} \bibinfo{year}{2023}\natexlab{}.
\newblock \showarticletitle{Model-free tracking control of complex dynamical
  trajectories with machine learning}.
\newblock \bibinfo{journal}{\emph{Nature Communications}} \bibinfo{volume}{14},
  \bibinfo{number}{1} (\bibinfo{year}{2023}), \bibinfo{pages}{5698}.
\newblock


\bibitem[Zhang et~al\mbox{.}(2020b)]%
        {zhang2020physics1}
\bibfield{author}{\bibinfo{person}{Ruiyang Zhang}, \bibinfo{person}{Yang Liu},
  {and} \bibinfo{person}{Hao Sun}.} \bibinfo{year}{2020}\natexlab{b}.
\newblock \showarticletitle{Physics-guided convolutional neural network
  (PhyCNN) for data-driven seismic response modeling}.
\newblock \bibinfo{journal}{\emph{Engineering Structures}}
  \bibinfo{volume}{215} (\bibinfo{year}{2020}), \bibinfo{pages}{110704}.
\newblock


\bibitem[Zhang et~al\mbox{.}(2020c)]%
        {zhang2020physics2}
\bibfield{author}{\bibinfo{person}{Ruiyang Zhang}, \bibinfo{person}{Yang Liu},
  {and} \bibinfo{person}{Hao Sun}.} \bibinfo{year}{2020}\natexlab{c}.
\newblock \showarticletitle{Physics-informed multi-LSTM networks for
  metamodeling of nonlinear structures}.
\newblock \bibinfo{journal}{\emph{Computer Methods in Applied Mechanics and
  Engineering}}  \bibinfo{volume}{369} (\bibinfo{year}{2020}),
  \bibinfo{pages}{113226}.
\newblock


\bibitem[Zhang et~al\mbox{.}(2023)]%
        {zhang2023skilful}
\bibfield{author}{\bibinfo{person}{Yuchen Zhang}, \bibinfo{person}{Mingsheng
  Long}, \bibinfo{person}{Kaiyuan Chen}, \bibinfo{person}{Lanxiang Xing},
  \bibinfo{person}{Ronghua Jin}, \bibinfo{person}{Michael~I Jordan}, {and}
  \bibinfo{person}{Jianmin Wang}.} \bibinfo{year}{2023}\natexlab{}.
\newblock \showarticletitle{Skilful nowcasting of extreme precipitation with
  {NowcastNet}}.
\newblock \bibinfo{journal}{\emph{Nature}} \bibinfo{volume}{619},
  \bibinfo{number}{7970} (\bibinfo{year}{2023}), \bibinfo{pages}{526--532}.
\newblock


\bibitem[Zhang et~al\mbox{.}(2020a)]%
        {zhang2020deep}
\bibfield{author}{\bibinfo{person}{Ziwei Zhang}, \bibinfo{person}{Peng Cui},
  {and} \bibinfo{person}{Wenwu Zhu}.} \bibinfo{year}{2020}\natexlab{a}.
\newblock \showarticletitle{Deep learning on graphs: A survey}.
\newblock \bibinfo{journal}{\emph{IEEE Transactions on Knowledge and Data
  Engineering}} \bibinfo{volume}{34}, \bibinfo{number}{1}
  (\bibinfo{year}{2020}), \bibinfo{pages}{249--270}.
\newblock


\end{thebibliography}


\appendix

\renewcommand{\thefigure}{S\arabic{figure}}
\setcounter{figure}{0} 

\renewcommand{\theequation}{S\arabic{equation}}
\setcounter{equation}{0} 

\renewcommand{\thetable}{S\arabic{table}}
\setcounter{table}{0} 


\section{Conservative Property of CiGNN}\label{si:conservative_property_of_CiGNN}

Despite the complex behavior exhibited in spatiotemporal dynamic systems, conservation laws provide a means to simplify the analysis of these systems. 
For instance, in fluid dynamics, the conservation of mass (also named continuity equation) is the key to the problem-solving.
Our proposed graph approximator is inspired by this conservation principle linking the \textit{flux} of a vector field through a closed surface to its divergence within the enclosed volume.
Suppose volume $\mathcal{V}$ is a compact subset of $\mathbb{R}^{3}$ and $\mathcal{S}$ (also indicated with $\partial \mathcal{V}$) is its piecewise smooth boundary, we have the detailed description shown as follows.

\textit{\textbf{Theorem 1:}
If there is a continuously differentiable tensor field $\mathbf{F}$ on a neighborhood of the volume $\mathcal{V}$, then we have that a volume integral over the volume $\mathcal{V}$ is equal to the surface integral over the boundary of the volume $\mathcal{V}$:}
\begin{equation}
\label{eq:theorem divergence theorem}
\iiint_{\mathcal{V}}(\nabla \cdot \mathbf{F})d \mathcal{V} = \iint\limits_{\mathcal{S}}(\mathbf{F} \cdot \mathbf {\hat {n}})d \mathcal{S},
\end{equation}
where $\mathbf {\hat {n}}$ is the outward-pointing unit normal.
The conservation principle (e.g., the divergence-free condition) formulated in Theorem 1 implies that while the flow quantity within a local region changes over time, the global net flow in and out remains nearly constant.
For further explanation, within the graph $G$, we denote ${flow}_{ij}$ as the flow that leaves from the node $i$, and ${flow}_{ji}$ the flow that enters to the node $i$. 
In a scalar or vector field, we consider that ${flow}_{ij} \geq {0}$ and ${flow}_{ji} \leq {0}$.
Here, a key assumption underlying the physics law is that ${flow}_{ij} = - {flow}_{ji}$, for $(i,j) \in G$.
Then, the conservation principle on the edge, or the divergence measurement on one vertex reads
\begin{equation}
\label{eq:conservation law on the edge}
{flux}_i = \sum_{(i,\cdot) \in G} {flow}_{i \cdot} + \sum_{(\cdot,i) \in G} {flow}_{\cdot i} ,
\end{equation}
where $flux_{i}$ is the total flow quantity through  the node $i$; $flow_{i\cdot}$ and $flow_{\cdot i}$ the flows that leave and enter the node $i$.
Subsequently, an easy corollary of Theorem 1 is described as follows.

\textit{\textbf{Corollary 1:}
If the volume ${\mathcal{V}}$ can be partitioned into some separate parts $\mathcal{V}_1, \mathcal{V}_2, \dots, \mathcal{V}_n$, the flux out $\Phi(\mathcal{V})$ of the original volume ${\mathcal{V}}$ should be equal to the sum of flux through each component volume:}
\begin{equation}
\label{eq:corollary divergence theorem}
\Phi \left(\mathcal{V} \right) =  \Phi \left(\mathcal{V}_1 \right)+ \Phi \left(\mathcal{V}_2 \right)+\dots+ \Phi \left(\mathcal{V}_n \right).
\end{equation}

Given the above corollary, for a large domain, we can divide it into multiple subdomains for separate processing.
Thus, we extend this theorem to the graph for learning spatiotemporal dynamics (e.g., PDEs) due to the essentially existence of subdomains of graph itself.
Let $G=(V,E,w)$ be a undirected graph, where $V$ is a set of nodes, $E$ a set of edges connecting the nodes $V$, and $w: E \rightarrow \mathbb{R}^{1}$ the edge weight function.
For simplicity, $w({u}_{i},{u}_{j})$ is expressed by $w_{ij}$ and $w({u}_{j},{u}_{i})$ is expressed by $w_{ji}$.

\textit{\textbf{Definition 1:}
With a vertex function $f \in \mathcal{H}(V)$ along the edge ${e}_{ij} \in E $, the weighted graph derivative (or weighted difference) of ${u}_{i}$ is expressed as:}
\begin{equation}
\label{eq:weighted graph derivative}
\partial_{{u}_{j}} f \left({u}_{i} \right) := \sqrt{w_{ij}} \left(f \left({u}_{j} \right)-f \left({u}_{i} \right) \right ),
\end{equation}
where the edge weight function $w_{ij}$ associates to every edge ${e}_{ij}$ a value.
Similarly, $w_{ji}$ also associates to every edge ${e}_{ji}$ a value.

\textit{\textbf{Definition 2:}
Based on the weighted graph derivative, we define the linear weighted gradient operator $\nabla_{w}:\mathcal{H}(V) \rightarrow \mathcal{H}(E)$ as:}
\begin{equation}
\label{eq:weighted gradient operator}
(\nabla_{w}f)({u}_{i},{u}_{j}) = \partial_{{u}_{j}} f({u}_{i}).
\end{equation}

\textit{\textbf{Definition 3:} 
The adjoint operator $\nabla^{*}_{w}$ of a edge function $F \in \mathcal{H}(E)$ at a vertex ${u}_{i} \in V$ has the following form:}
\begin{equation}
\label{eq:another adjoint operator}
(\nabla^{*}_{w} F)({u}_i) = \frac{1}{2} \sum_{j \in \mathcal{N}_{i}} \sqrt{w_{ij}} (F_{ji}-F_{ij}),
\end{equation}
where $\mathcal{N}_{i}$ denotes the neighborhood node's index set for node $i$.
For simplicity, we simplified $F({u}_{i},{u}_{j})$ and $F({u}_{j},{u}_{i})$ as $F_{ij}$ and $F_{ji}$.

\textit{\textbf{Theorem 2:}
For all $f \in \mathcal{H}(V), F \in \mathcal{H}(E)$, the linear adjoint operator $\nabla^{*}_{w}:\mathcal{H}(E) \rightarrow \mathcal{H}(V)$ of the weighted gradient operator should satisfy the following condition:}
\begin{equation}
\label{eq:weighted adjoint operator}
<\nabla_{w}f, F>_{\mathcal{H}(E)} = <f, \nabla^{*}_{w} F>_{\mathcal{H}(V)}.
\end{equation}

\textit{\textbf{Definition 4:}
Using the linear adjoint operator, the weighted divergence operator on graphs can be expressed as:}
\begin{equation}
\label{eq:weighted divergence operator}
div_{w} := - \nabla^{*}_{w}.
\end{equation}

With above definition, we immediately get the following inference that the divergence on a graph can be interpreted as the $flux$ of the edge function in each vertex of the graph. Then we rewrite the original PDE in Eq. \eref{eq:continuity} into the graph form as follows
\begin{equation}
\label{eq:conservation form}
\frac{\partial {u}_i}{\partial t} + div_w(F({u}_i)) = {s}({u}_{i}),
\end{equation}
where ${s}(\cdot)$ represents the source term.
Then, substituting Eqs.~\eref{eq:another adjoint operator} and \eref{eq:weighted divergence operator} into Eq.~\eref{eq:conservation form}, we get the following equation:
\begin{equation}
\label{eq:graph divergence operator on undirected graph}
\frac{\partial {u}_i}{\partial t} = \frac{1}{2} \sum_{j \in \mathcal{N}_{i}} \sqrt{w_{ij}} \left(F_{ji}-F_{ij}  \right) + {s}({u}_{i}).
\end{equation}

It is noted that in a directed graph, $w_{ij}$ may not be equal to $w_{ji}$, which aligns with the actual situation that there exist conservation and inequality terms which govern the dynamics.
Hence, we re-write Eq.~\eref{eq:graph divergence operator on undirected graph} and obtain the following equation on a directed graph:
\begin{equation}
\label{eq:graph divergence operator on directed graph}
\frac{\partial {u}_i}{\partial t} = \frac{1}{2} \sum_{j \in \mathcal{N}_{i}}  \left(\sqrt{w_{ji}}F_{ji}-\sqrt{w_{ij}}F_{ij} \right) + {s}({u}_{i}).
\end{equation}
We realize that the conservation term can be formulated by the skew-symmetric tensor and the inequality by the symmetric tensor \cite{tadmor1984skew}. Instead of constructing such tensors directly, we implicitly build a learnable scheme to retain the above structure property:
\begin{equation}
\label{eq:CiGNN}
    \frac{\partial {u}_i}{\partial t} = 
    \sum_{j \in \mathcal{N}_{i}}  
    \Big(
    \underbrace{ F_{ji}-F_{ij}}_{\text{skew-symmetric}} 
    + 
    \underbrace{ \phi\left(F_{ji}+F_{ij}\right)}_{\text{symmetric}} 
    \Big),
\end{equation}
where $\phi(\cdot)$ is a differential function to model the inequality and, meanwhile, learn the unknown source term (e.g., entropy inequality). Here, we omit the coefficient of 1/2 for convenience.
Here, the skew-symmetric and symmetric features on the right-hand side of Eq.~\eref{eq:CiGNN} are constructed to approximate the conservation and inequality principles, respectively.
In particular, we use the addition operation rather concatenation for the symmetric part in Eq.~\eref{eq:CiGNN} to guarantee the invariance and interchangeability.

Note that the above derivation is done in the physical space, which inspires us to design a learnable model (e.g., CiGNN) in the latent space following a similar setting. The graph network is not restricted to highly regular structures, which, instead, can be applied to represent abstract relationships on any irregular meshes.

\section{Dataset Generation}\label{si:dataset_generation}

\subsection{2D Burgers Equation}

In our work, we consider the Burgers equation within a 2D space setting, which is given by
\begin{subequations}\label{eq:burgers}
\begin{align}
\frac{\partial {u}(x,y,t\geq0)}{\partial t} &= \nu \nabla^2{u} - {u} \frac{\partial {u}}{\partial x} - {v} \frac{\partial {u}}{\partial y},\\
\frac{\partial {v}(x,y,t\geq0)}{\partial t} &= \nu \nabla^2{v} - {u} \frac{\partial {v}}{\partial x} - {v} \frac{\partial {v}}{\partial y}.
\end{align}
\end{subequations}

Here $\nu>0$ represents the viscous property of the fluid, and $u(x,y,t)$ and $v(x,y,t)$ denote the velocity components at the indicated spatial and temporal coordinates. Specifically, we leverage periodic boundary conditions to generate the simulation data within a spatial domain of $\Omega = (0,1)^{2}$ and a time duration of $t \in[0, 1]$. The viscous coefficient $\nu$ is set to $0.01$. Additionally, the initial conditions ${u}_0(x,y)={u}(x,y,t = 0)$ and ${v}_0(x,y)={v}(x,y,t = 0)$ for each component are generated with:
\begin{subequations}
\begin{align}
\tilde{u}_0(x, y) &= \sum_{a_i=0}^{n_a}\sum_{b_j=0}^{n_b}{\lambda_{u}\cos{(\iota)} + \gamma_{v}\sin{(\iota)}}, \\
u_0(x, y) &= \frac{1}{3} \times \left( 2 \times \frac{\tilde{u}_0(x, y)}{\max{\tilde{u}_0(x, y)}} + c_u \right),\\
\tilde{v}_0(x, y) &= \sum_{a_i=0}^{n_a}\sum_{b_j=0}^{n_b}{\lambda_{u}\cos{(\iota)} + \gamma_{v}\sin{(\iota)}}, \\
v_0(x, y) &= \frac{1}{3} \times \left( 2 \times \frac{\tilde{v}_0(x, y)}{\max{\tilde{v}_0(x, y)}} + c_v \right),\\
\iota &= 2 \pi ((a_i-\frac{n_a}{2})x+(b_j-\frac{n_b}{2})y),
\end{align}
\end{subequations}
where $N=10, \delta t = 0.001, \delta x = \delta y = 1/50$, $\lambda_{u},\gamma_{u},\lambda_{v},\gamma_{v} \sim \mathcal{N}(0, 1)$, and ${c_u,c_v} \sim \mathcal{N}(-1, 1)$. The ground-truth data is calculated using a $4$th-order Runge–Kutta time integration method~\cite{ren2022phycrnet}.

\subsection{3D GS RD Equation}

The 3D GS RD equation is expressed by:
\begin{subequations}\label{eq:gs}
\begin{align}
\frac{\partial {u(x,y,z,t\geq0)}}{\partial t} &= D_u\nabla^2 {u} - {u}  {v}^{2} + \alpha  (1- {u}),\\
\frac{\partial {v(x,y,z,t\geq0)}}{\partial t} &= D_v\nabla^2{v} +{u} {v}^{2}-(\beta+\alpha) {v}.
\end{align}
\end{subequations}
where $u(x,y,z,t)$ and $v(x,y,z,t)$ are the velocities.
$D_u$ and $D_v$ are the corresponding diffusion coefficients for $u$ and $v$, respectively.
$\beta$ is the conversion rate, $\alpha$ is the in-flow rate of $u$ from the outside, and $(\alpha + \beta)$ is the removal rate of $v$ from the reaction field.

The simulation datasets are obtained by solving the following initial-boundary value problem on $\Omega = [0, 96]^{3}$ with periodic boundary conditions. Moreover, we define $D_u = 0.2, D_v = 0.1, \alpha = 0.025, \beta = 0.055, \delta t = 0.25, \delta x = 2$. 
The initial conditions ${u}_0(x,y,z)=u(x,y,z,t = 0)$ and ${v}_0(x,y,z)=v(x,y,z,t = 0)$ for each component are generated as follows:
\begin{subequations}
\begin{align}
u_0(x, y ,z) &= (1-rt) \times {1} + rt \times {\lambda_u},  \\
v_0(x, y ,z) &= {0} + rt \times {\lambda_v},
\end{align}
\end{subequations}
where $rt = 0.1$ and ${\lambda_u}, {\lambda_v} \sim \mathcal{N}(0, 1)$.
The $4$th-order Runge–Kutta time integration method is also applied.

\subsection{2D CF Dataset}

The CF dataset simulates the temporal evolution of incompressible flow past a long cylinder. It poses challenges to dynamics prediction due to the occurrence of periodic flow patterns. In this paper, we simulate CF datasets with various positions of the cylinder and different sizes of the diameter of cylinders using the {COMSOL} software \footnote{\url{https://cn.comsol.com/model/flow-past-a-cylinder-97}}. 
We define the spatial domain as $ x \in [0, 1.2] [m] $ and $ y \in [0, 0.41] [m]$ and the time duration as  $t \in [0,10] [s]$.
In addition, we generate CF examples with the no-slip boundary condition.

Furthermore, we fix the median initial velocity $U_{mean} = 1~[\text{m/s}]$ and viscosity $\nu= 10^{-3}~[\text{Pa} \cdot \text{s}]$, but modify the radius sizes $r$ and the positions $(c_x, c_y)$ of cylinders. To be more specific, we sample these two parameters uniformly where $ r \in [ 0.04, 0.081 ], C \in [0.1, 0.31] \times [0.1, 0.31]$ (unit: $[\text{m}]$).The Reynolds numbers $Re$ are within $[800, 1600]$.

\subsection{2D Real-world Black Sea Dataset}

The Black Sea (BS) dataset is obtained using the Nucleus for European Modeling of the Ocean (NEMO) ~\footnote{ \url{https://doi.org/10.5281/zenodo.1464816}} general circulation ocean modelwithin the Black Sea domain. The NEMO model is driven by atmospheric surface fluxes computed through bulk formulation, utilizing ECMWF ERA5~\cite{hersbach2020era5} atmospheric fields with a spatial resolution of $0.25^{\circ}$ and a temporal resolution of 1 hour. Additionally, the NEMO model is online coupled to the OceanVar assimilation scheme~\cite{FARINA2015631}, allowing for the assimilation of sea level anomaly along-track observations from Copernicus Marine Environment Monitoring Service~(CMEMS)
~\footnote{ \url{https://marine.copernicus.eu/}} 
datasets. In our paper, we preprocess the spatial information by uniformly sampling 1,000 points from the total computational domain including over 40,000 real measurement points for model learning and prediction. Furthermore, we divide the data into annual units and resample the dataset with one snapshot per day to enable effective learning of periodic information in time series data. The training dataset spans from June 1, 1993, to December 31, 2017, the validation dataset covers January 1, 2018, to December 31, 2020, and the test dataset comprises data from January 1, 2021, to June 31, 2021.

\end{document}